\documentclass[a4paper,fleqn]{cas-dc}

\usepackage[numbers,sort]{natbib}
\usepackage{multirow}

\def\tsc#1{\csdef{#1}{\textsc{\lowercase{#1}}\xspace}}
\tsc{WGM}
\tsc{QE}

\begin{document}
\let\WriteBookmarks\relax
\def\floatpagepagefraction{1}
\def\textpagefraction{.001}

\shorttitle{PatTree}    

\shortauthors{J. Gehrmann et al.}  

\title[mode = title]{PatTree: a novel approach for automated creation of multimodal, graph-based patient representations for medical classification tasks}



%

\author[1]{Julia Gehrmann}[orcid=0000-0002-4101-545, linkedin=julia-gehrmann-b69b1a1b3/]
\cormark[1]
\ead{julia.gehrmann1@uk-koeln.de}
\ead[url]{https://bi-koeln.de/en/team/julia-gehrmann}
\credit{Conceptualization, Data curation, Formal analysis, Investigation, Methodology, Project administration, Software, Validation, Visualization, Writing – original draft, Writing – review and editing}

\author[1]{Lars Quakulinski}[orcid=0009-0009-2475-3326, linkedin=lars-quakulinski/]
\ead{lars.quakulinski@rwth-aachen.de}
\ead[url]{https://bi-koeln.de/en/team/lars-quakulinski}
\credit{Conceptualization, Data curation, Formal analysis, Investigation, Methodology, Software, Validation, Writing – review and editing}

\author[1]{Hamza Naseem}[orcid=0009-0002-9249-9082, linkedin=hamza-naseem-4a4a912a5/]
\ead{hamza.naseem@uk-koeln.de}
\ead[url]{https://bi-koeln.de/en/team/hamza-naseem}
\credit{Formal analysis, Investigation, Methodology, Software, Validation, Writing – review and editing}

\author[1,2,3]{Oya Beyan}[orcid=0000-0001-7611-3501,linkedin=oya-deniz-beyan-50a02a8/]
\ead{oya.beyan@uni-koeln.de}
\ead[url]{https://oyabeyan.info/}
\credit{Conceptualization, Project administration, Resources, Supervision, Writing – review and editing}

\author[]{for the Alzheimer's Disease Neuroimaging Initiative}[]
\fnmark[1]

\affiliation[1]{organization={Institute for Biomedical Informatics, University of Cologne, Medical Faculty and University Hospital Cologne},
            addressline={Kerpener Straße 62}, 
            city={Cologne},
            postcode={50937}, 
            country={Germany}}

\affiliation[2]{organization={Division of Computer Science, Department of Mathematics and Computer Science, Faculty of Mathematics and Natural Sciences, University of Cologne},
            addressline={Albertus-Magnus-Platz}, 
            city={Cologne},
            postcode={50923}, 
            country={Germany}}

\affiliation[3]{organization={Department of Data Science and Artificial Intelligence, Fraunhofer Institute for Applied Information Technology},
            addressline={Schloss Birlinghoven}, 
            city={Sankt Augustin},
            postcode={53757}, 
            country={Germany}}


\cortext[1]{Corresponding author}

\fntext[1]{Data used in preparation of this article were obtained from the Alzheimer's Disease Neuroimaging Initiative (ADNI) database (adni.loni.usc.edu). As such, the investigators within the ADNI contributed to the design and implementation of ADNI and/or provided data but did not participate in the analysis or writing of this report. A complete listing of ADNI investigators can be found at: \url{http://adni.loni.usc.edu/wp-content/uploads/how_to_apply/ADNI_Acknowledgement_List.pdf"}}


\begin{abstract}
Access to holistic, multimodal data improves the performance of Artificial Intelligence (AI) in medical classification tasks compared to utilizing single modalities or data sources. However, the inherent heterogeneity and complexity of clinical real-world data pose significant challenges to structured data analysis and AI application. This heterogeneity includes missing values, multiple time points, diverse modalities, and inconsistent formats and semantics. Data harmonization prior to data integration tackles this challenge but remains resource-intensive and error-prone, limiting the scalability and reproducibility of holistic, AI-driven decision support on clinical real-world data. We therefore propose PatTree, a graph-based, holistic representation of patients that can be derived from real-world clinical data through the automated structuring of multimodal clinical data. PatTree enables early-stage data integration without relying on pre-standardized inputs. While representing heterogeneous clinical data within a unified knowledge graph, PatTree preserves the semantic relationships between data elements across modalities and data sources, facilitating interoperability and machine-interpretable data access. Using a subset of the ADNI-1 cohort (n = 763), we demonstrate that classification of patients is directly feasible on PatTree reaching state-of-the-art classification performance. In the three-class classification task distinguishing Alzheimer's disease, mild cognitive impairment, and cognitively normal individuals, we achieve a balanced accuracy of 98.5\% and an F$_1$ score of 0.987 on the held-out test set. Our results show that assumption-free, automated structuring of multimodal medical data can serve as a scalable foundation for clinical AI pipelines bypassing tedious data preparation and standardization.
\end{abstract}


\begin{highlights}
\item automated integration of multimodal, longitudinal patient data 
\item semantic enrichment of measured values
\item tree-based patient representation
\end{highlights}


\begin{keywords}
Multimodal Data \sep Knowledge Graph \sep Medicine \sep Patient Representation \sep Classification
\end{keywords}

\maketitle

\section{Introduction}\label{intro}
Clinical real-world data is inherently heterogeneous at multiple levels. At the most fundamental level, each patient follows a unique clinical journey with certain measurements and examinations, resulting in different sets of available observations and data elements across patients. Beyond this, the data itself varies in modality, format, and recording frequency within and across clinical cases. Finally, the underlying clinical information systems add another layer of structural heterogeneity through their customized database schemas, storage conventions, and institutional differences in what information is captured and where. This multi-level heterogeneity challenges systematic analyses on clinical real-world data motivating the harmonization and integration of such data \cite{gehrmann2023prevents}.

On the other hand, there are inherent structural and medically substantiated relationships between clinical data elements that reflect hidden graph structures. Prior work on graphs and Graph Neural Networks (GNNs) shows that leveraging such graph structures and, in particular, their topological information increases analysis and prediction performance compared to analyzing the data elements in tabular format \cite{gori2005new, ali2025graph}. This fact makes the medical domain a promising application field for GNNs. The missing link, however, is uncovering the hidden graph structures and turning heterogeneous clinical real-world data into medical Knowledge Graphs (KGs) explicitly representing relationships between data elements. Prior work in this field focuses on integrating clinical data into graph structures leveraging data standards, ontologies or existing KGs. One such approach is harmonizing the data and, subsequently, utilizing the structure of the harmonized data for KG creation. The harmonization is typically achieved by converting the data into Fast Healthcare Interoperability Resources (FHIR) or the Observational Medical Outcomes Partnership (OMOP) Common Data Model \cite{xiao2022fhir, mazein2024medax, chytas2024mapping, kang2024evolution}. Another approach is mapping the clinical data directly to existing graph structures through named entity recognition (NER) i.e. mapping clinical entities in the source data to concepts from an ontology or entities in an existing KG \cite{frau2025connecting, sachdeva2022using}. However, both approaches have significant bottlenecks limiting their applicability: Data harmonization is a time-consuming and error-prone process \cite{le2020challenges}, while NER is an open research field itself representing a potential source for errors through mismatches and missed matches. NER often fails to capture temporal or subdomain-specific concepts and aspects \cite{goyal2025named}.

To address these gaps, we propose PatTree, a multimodal, graph-based patient representation for medical classification tasks created automatically and assumption-free from heterogeneous medical data records. The PatTree approach is assumption-free in that it does not require data harmonization or NER but leverages the natural structure of medical data given by patient journeys and the architecture of clinical information systems. Based on these structures, which are always available, the PatTree approach creates a holistic patient representation that can, subsequently, be mined by GNNs, for instance, for classification tasks.

\section{Related Work}\label{rl}

\subsection{Data Representations in Medicine and Medical Research}\label{rl.data_representation}
Clinical information systems typically store patient data in relational databases, where tables and their relationships mirror the day-to-day operations of hospitals and clinics \cite{overhage2012validation}. For research and data exchange, however, this raw relational data is usually transformed into standardized formats. Common Data Models (CDMs) such as OMOP harmonize heterogeneous source data into a shared relational schema, enabling federated, reproducible analyses across institutions \cite{hripcsak2015observational, klann2018web}. Alongside CDMs, interoperability standards like HL7 FHIR represent clinical data as structured, resource-based JSON documents, prioritizing exchange between systems over analytical convenience \cite{ayaz2021fast}. A recent systematic comparison shows that OMOP and FHIR consistently outperform competing CDMs and standards, yet also finds that no single representation dominates across all use cases, underscoring the need for transformation and interoperability between formats \cite{finster2025common}. Beyond structured and semi-structured formats, medical data increasingly includes complex modalities such as imaging, video, and omics data, further complicating the choice of a suitable representation \cite{sun2023scoping}. In contrast to the table- or document-centric views imposed by relational databases, OMOP, and FHIR, graph-based representations have gained popularity in research because they can natively encode the rich, heterogeneous relationships between patients, clinical concepts, and events \cite{choi2017gram, xiao2022fhir, mazein2024medax, chytas2024mapping, kang2024evolution}. Rather than flattening clinical knowledge into rows or nested resources, graphs represent diagnoses, medications, procedures, and patients as nodes connected by clinically meaningful edges, preserving structural information that would otherwise be lost or require costly joins \cite{choi2020learning, abuhantash2024comorbidity}. This structural expressiveness makes graph-based representations particularly well-suited for AI-based analysis, and GNNs have consequently become an established methodological family alongside transformer- and autoencoder-based approaches for downstream knowledge discovery and clinical decision support (CDS) \cite{zheng2025scoping}.

\subsection{Multimodal Analysis in Medical Research}\label{rl.multmod}
During diagnosis and treatment, medical patients typically undergo various examinations, generating data in diverse formats such as images, videos, single numbers, free text, or omics data. These data formats are also referred to as modalities \cite{lee2017medical, behrad2022overview}. Each modality offers a unique perspective on the medical case and clinical decisions rely on a comprehensive, joint consideration of all relevant data modalities.
In contrast to clinical decision processes, traditional medical research often focuses on individual data modalities. Such unimodal approaches do not suffice to model the complex processes in the human body and their interactions \cite{kline2022multimodal}. To enable more holistic insights, researchers have begun exploring multimodal data integration combining data from different modalities. Existing approaches typically integrate the modalities at the result stage. This means that the analyses are performed for each modality individually and, afterwards, the analysis results are integrated. This approach is referred to as late integration. While it offers some benefits like easy implementation, late integration overlooks the complex relationships between different modalities \cite{bokade2021cross}. However, such intermodal relationships hold valuable information that can be exploited to improve the analysis accuracy and, thus, medical knowledge discovery and CDS. Intermodal relationships can be incorporated in the integrated multimodal data, if the modalities are combined prior to the analysis. This approach is referred to as early integration. Early integration nowadays often applies AI to fuse the modalities. Moreover, the subsequent downstream analysis aiming for knowledge discovery or CDS most commonly uses AI-based approaches \cite{boehm2022harnessing}.

\subsection{Image Feature Extraction}\label{rl.img_feat_extr}
Feature extraction is a fundamental step in image-based machine learning pipelines, retrieving compact, informative vector representations from raw pixel data. Early approaches relied on hand-crafted features designed by domain experts to capture specific visual properties. These features encode local texture, edge, and gradient information in a computationally efficient manner. While these methods offer interpretability and low computational cost, their representational capacity is inherently limited by the assumptions embedded in their design \cite{hallur2025image, dehbozorgi2025comparative}.

Advances in AI research towards deep learning (DL) transformed the field of feature extraction. Convolutional Neural Networks (CNNs) can learn hierarchical, task-relevant feature representations directly from raw image data through end-to-end training. Early convolutional layers capture low-level patterns such as edges and textures, while deeper layers encode increasingly abstract and semantically meaningful features. In classification settings, the activations of the final fully connected layer serve as a compact image embedding, encoding the information most discriminative for the task at hand. Compared to hand-crafted features, these DL-based features tend to be less interpretable as they typically result from an opaque feature extraction process. \cite{dehbozorgi2025comparative, hallur2025image, chaki2026deep}

The application of feature extraction to brain imaging introduces domain-specific considerations, as neuroimaging modalities such as structural magnetic resonance imaging (sMRI) and functional MRI (fMRI) produce high-dimensional volumetric data with rich structural and functional information. A well-established paradigm in this domain is radiomics, which extracts quantitative features from medical images \cite{bevilacqua2023radiomics,mayerhoefer2020introduction}. Radiomics features are broadly categorized into first-order statistics such as mean intensity, variance, skewness, and kurtosis that are derived directly from the voxel intensity distribution, as well as morphological descriptors including region volumes, surface area, and compactness. Higher-order texture features, such as those derived from Gray-Level Co-occurrence Matrices (GLCM) or Laplacian of Gaussian (LoG) filters, capture spatial relationships between voxels and have demonstrated utility in characterizing tissue heterogeneity \cite{mayerhoefer2020introduction}.

Beyond radiomics, DL methods have increasingly been applied to neuroimaging for automated feature extraction. CNNs adapted for three-dimensional volumetric data, such as 3D-CNNs and U-Net variants, have shown strong performance on tasks including brain tumor segmentation, lesion detection, and disease classification \cite{chaki2026deep}. These models implicitly learn feature representations tailored to the specific diagnostic task, often outperforming classical radiomics features when sufficient training data are available \cite{dehbozorgi2025comparative}. Hybrid approaches combining radiomics with deep features have also been explored to leverage the complementary strengths of domain knowledge and learned representations \cite{hallur2025image}.

\subsection{Natural Language Embedding}\label{rl.nat_lang_emb}
Natural Language Embedding translates natural language such as words or sentences into numerical vectors of fixed dimensionality, the embeddings. This translation makes linear algebra, statistics, and machine learning directly applicable to text. Early approaches to natural language embedding include TF-IDF weighting and the Bag of Words (BoW) model, which represent words and documents as numerical scores or sparse vectors based on word occurrence frequencies \cite{zhang2024comparative}. While embedding natural language into vector representations, these models do not capture meaning, i.e. semantically similar words are treated as entirely unrelated. Therefore, researchers began developing more expressive representations, aiming to encode semantic similarity: words with related meanings would be assigned similar vectors. This shift was marked by the introduction of the influential Word2Vec model in 2013, whose architecture learns dense word embeddings by predicting the context surrounding a given word \cite{mikolov2013efficient}. State-of-the-art performance is achieved with transformer models that make increasingly efficient use of the attention mechanism introduced in 2017 \cite{vaswani2017attention, salem2025transformer}. At its core, the attention mechanism allows a model to dynamically weigh the relevance of different elements in an input sequence when computing a representation for a given element, rather than relying on a fixed context window or a single compressed summary vector \cite{vaswani2017attention}. Nowadays, there exist a variety of pretrained transformer models differing in their input (words or sentences), the domain they are trained in (e.g. general, biomedical) and the dimension of the embeddings they produce \cite{salem2025transformer}.

\subsection{Knowledge Graphs in the Medical Domain}\label{rl.kg_med}
A KG is a directed edge-labeled graph, i.e. a tuple $G = (V, E, L)$ of nodes $V$, edges $E$ and edge labels $L$ which explicitly models some kind of knowledge. The nodes $V$ represent entities; the edges $E$ link represent relationships between entities; and the edge labels $L$ specify the kind of relationship \cite{hogan2022knowledge}.

Prior work shows that KGs can, in particular, model biomedical and clinical knowledge by linking respective data meaningfully. Such biomedical KGs enable precision medicine by coupling related data and information spread across databases. Depending on the primary purpose of the KG, nodes represent e.g. diagnoses, symptoms, proteins, medication, or measurements. The represented information and data typically come from publicly available databases, guidelines, or EHRs \cite{morris2023scalable, chandak2023building, aldughayfiq2023capturing, guluzade2021demographic, bloor2023towards, seneviratne2021personal, li2020real, sengupta2025medaka, yang2025large, abuhantash2024comorbidity}. Edges are inferred from existing ontologies \cite{morris2023scalable, chandak2023building, aldughayfiq2023capturing, guluzade2021demographic, bloor2023towards, seneviratne2021personal}, defined from scratch \cite{li2020real} or retrieved by use of LLMs \cite{sengupta2025medaka, yang2025large}.

Since the underlying data sources for biomedical KG creation are often structured, several approaches have been proposed to construct KGs directly from such structured data. These approaches include FHIR-based methods, such as FHIR RDF \cite{xiao2022fhir} or MeDAX \cite{mazein2024medax}, as well as OMOP-to-Graph pipelines \cite{chytas2024mapping,kang2024evolution}. However, these approaches share common limitations. Firstly, they require the source data to be harmonized into a data standard or model such as FHIR or OMOP prior to KG construction. Secondly, only including tabular data represented in the respective standard excludes important information from further modalities while KGs carry the potential of integrating multimodal data \cite{gehrmann2024early}.

\subsection{Knowledge Graph Classification}\label{rl.KG_class}
KG Classification refers to the process of assigning class labels to individual KGs. The traditional approach to KG classification firstly learns low-dimensional vector embeddings of nodes and edges and, subsequently, applies a classical machine learning or neural classifier on these representations. A fundamental limitation shared by such methods is their transductive nature, i.e. embeddings are learned for a fixed set of graphs and do not generalize to unseen entities or graphs without retraining. Furthermore, the embeddings do not capture the full topological information of the KG \cite{gori2005new, grohe2020word2vec}.

To address these shortcomings, GNNs have emerged as an increasingly used alternative, learning task-specific embeddings directly from the graph structure and classify them without a separate embedding step. GNNs operate through iterative message passing, wherein each node aggregates feature information from its local neighborhood and updates its representation accordingly. This mechanism enables the model to jointly capture both node-level information and relational topology, propagating information across multiple hops through successive layers. Eventually, each node carries its final node embedding \cite{gori2005new, grohe2020word2vec, morris2022graph}.

For graph classification, global readout functions such as sum, mean, or hierarchical pooling are applied over all node embeddings to produce a fixed-size graph embedding. A subsequent classification layer maps this embedding directly to a graph-level prediction \cite{morris2022graph}.
Compared to two-step pipelines, GNNs offer several key advantages: inductive generalization to unseen graphs, tighter integration between representation learning and classification task, and the ability to consider rich topological information \cite{gori2005new, morris2022graph}. 

The attention mechanism, already meantioned in section \ref{rl.nat_lang_emb}, also made its way into the GNN realm through Graph Attention Networks (GATs) in 2018. GATs learn adaptive, edge-specific weighting schemes rather than relying on fixed or uniform aggregation of node vectors. In the original formulation, GATs compute attention coefficients for each node's neighbors using a shared, learnable attention mechanism, allowing the model to assign different importance to different neighbors when aggregating messages, rather than treating all neighbors equally as in GCNs \cite{velivckovic2018graph}. However, subsequent work identified that the original GAT ranks neighbors in an order that is fixed for all query nodes and independent of the query node's own representation. As a result, the model's expressiveness is constrained. To address this, GATv2 introduces a simple but effective modification to the order of operations in the attention computation, replacing the original's fixed scoring function with a scoring function that allows for dynamic attention, in which the ranking of neighbor importance can vary depending on the query node. This enables GATv2 to better capture the neighborhood structure and has been shown to outperform the original GAT across a range of benchmark tasks, particularly on datasets requiring more complex node interactions \cite{brody2021attentive}.

\subsection{AI-based Alzheimer Diagnosis}\label{rl.AD_diag}
The high availability of large-scale neuroimaging and clinical datasets such as ADNI and OASIS, promoted research on AI-based diagnosis of Alzheimer's Disease (AD) significantly. Typically, AD diagnosis is formalized as a binary classification task discriminating AD from cognitive normal (CN) cases or as a multi-class classification task additionally considering mild cognitive impairment (MCI) as a third, intermediate class. Recent advances in AI-based AD diagnosis are largely implemented with DL methods, which enable the extraction of complex patterns from high-dimensional biomedical data \cite{shaikh2025deep}.

Unimodal classification methods rely on a single data source, most commonly neuroimaging modalities. Among these, sMRI and fMRI are used most frequently due to their high availability and strong representation of neurodegenerative changes. DL models, particularly CNNs, have shown strong performance in unimodal settings by learning hierarchical representations directly from raw imaging data. Reported accuracies for binary classification tasks (i.e. CN vs. AD) range from approximately 71\% to over 99\%, depending on modality and dataset \cite{ali2025graph, ebrahimi2021convolutional}. However, these figures must be interpreted with caution: a comprehensive evaluation of deep learning models for AD classification found that nearly half of published models contained data leakage due to incorrect subject-level data splitting, with correct patient-level partitioning reducing reported accuracy by up to 10 percentage points \cite{wen2020convolutional}. Under rigorously leakage-free conditions, binary CN vs. AD classification using sMRI deep learning yields accuracies of approximately 90-93\%\cite{zhou2025deep, hussain2025alzformer}. Despite these successes, unimodal approaches are inherently limited by their reliance on a single aspect of the disease. AD is multifactorial, involving structural, functional, molecular, and clinical changes, which cannot be fully captured by any single modality \cite{shaikh2025deep}. Furthermore, without further adaptation CNNs lack explainability, limiting their applicability in clinical practice.

To address these limitations, multimodal approaches integrate heterogeneous data sources, including combinations of neuroimaging (sMRI, fMRI, PET), clinical assessments, cognitive scores, genetic information, and cerebrospinal fluid (CSF) biomarkers. These methods exploit complementary information across modalities to better characterize disease pathology \cite{wang_2018-07_NovelMultimodalMRI, abuhantash2024comorbidity}.
Prior work shows that multimodal models generally outperform unimodal approaches. A recent review of 73 studies reported that multimodal approaches achieve higher and more stable classification accuracy across tasks, with top performances exceeding 99\% in accuracy for binary AD classification \cite{ali2025graph}. Similarly, multimodal systems combining imaging and non-imaging data have demonstrated diagnostic performance comparable to clinical experts in multi-stage classification settings \cite{qiu2022multimodal}.
Despite promising results, several challenges remain.

Multimodal models require large, well-aligned datasets, which are often limited by missing modalities and cohort heterogeneity \cite{shaikh2025deep}. Furthermore, in the three-class setting, the classes are not as easily separable due to the continuous nature of AD development from CN over MCI to AD. Early MCI resembles CN, while late MCI is difficult to distinguish from AD \cite{goryawala2015inclusion}. This is reflected in empirical results. Rigorous leakage-free studies on the three-class CN vs. MCI vs. AD problem report accuracies in the range of 59-98\%, depending on the study, ranging from significantly below to only slightly below binary CN vs. AD performance \cite{mieling2026predicting, diogo2022early, tanveer2020machine, pellegrini2018machine, abuhantash2024comorbidity}.

\section{Methods}\label{methods}
We propose PatTree, a graph-based, multimodal patient representation inherently handling heterogeneity of data across cohorts and leveraging intrinsic data structures. PatTree's basis is a tree in which the root node represents the patient, leaf nodes represent medical data features of the patient and internal nodes represent the relation of the patient to the data features inferred from basic metadata. Firstly, the data is preprocessed by feature extraction from images, normalization of continuous features, and semantic enrichment through embedding of string data and metadata. Subsequently, the graph is constructed leveraging structural information carried by the patient journey and the architecture of clinical information systems reflected in the metadata. Finally, the created PatTrees allow for direct patient classification through GAT-based graph classification.

\subsection{Data Preprocessing}\label{meth.preprocessing}
PatTree's data preprocessing prepares harmonization of heterogeneous, multimodal data into numerical vectors of fixed dimensionality which can later be used as node property vectors in the patient graph. To this end, we extracted features from images and strings as outlined in sections \ref{meth.img_feature_extr} and \ref{meth.name_and_term_emb}. Furthermore, we apply $z$-score normalization on all continuous features, i.e. all features with a numerical data type, to harmonize value ranges. The results of this preprocessing are later utilized for creating node property vectors in PatTree construction.

\subsubsection{Image Feature Extraction}\label{meth.img_feature_extr}
We compare two approaches for image feature extraction from MRI data: unsupervised and supervised features, i.e. features determined without and with knowing the patients' class, respectively. The first, unsupervised approach is performing brain segmentation into brain regions of interest (ROI) to calculate ROI volumes as radiomics features. The second, supervised approach utilizes a CNN to learn prototypes, i.e. prototypical image patches from the training data, and determines similarities of MRI scans to these learned prototypes. The ROI volumes and prototype similarities can then be interpreted as features representing the MRI scans.

\subsubsection{Embedding of Node Names and Category Terms}\label{meth.name_and_term_emb}
\begin{figure*}
    \centering
    \includegraphics[width=0.9\textwidth]{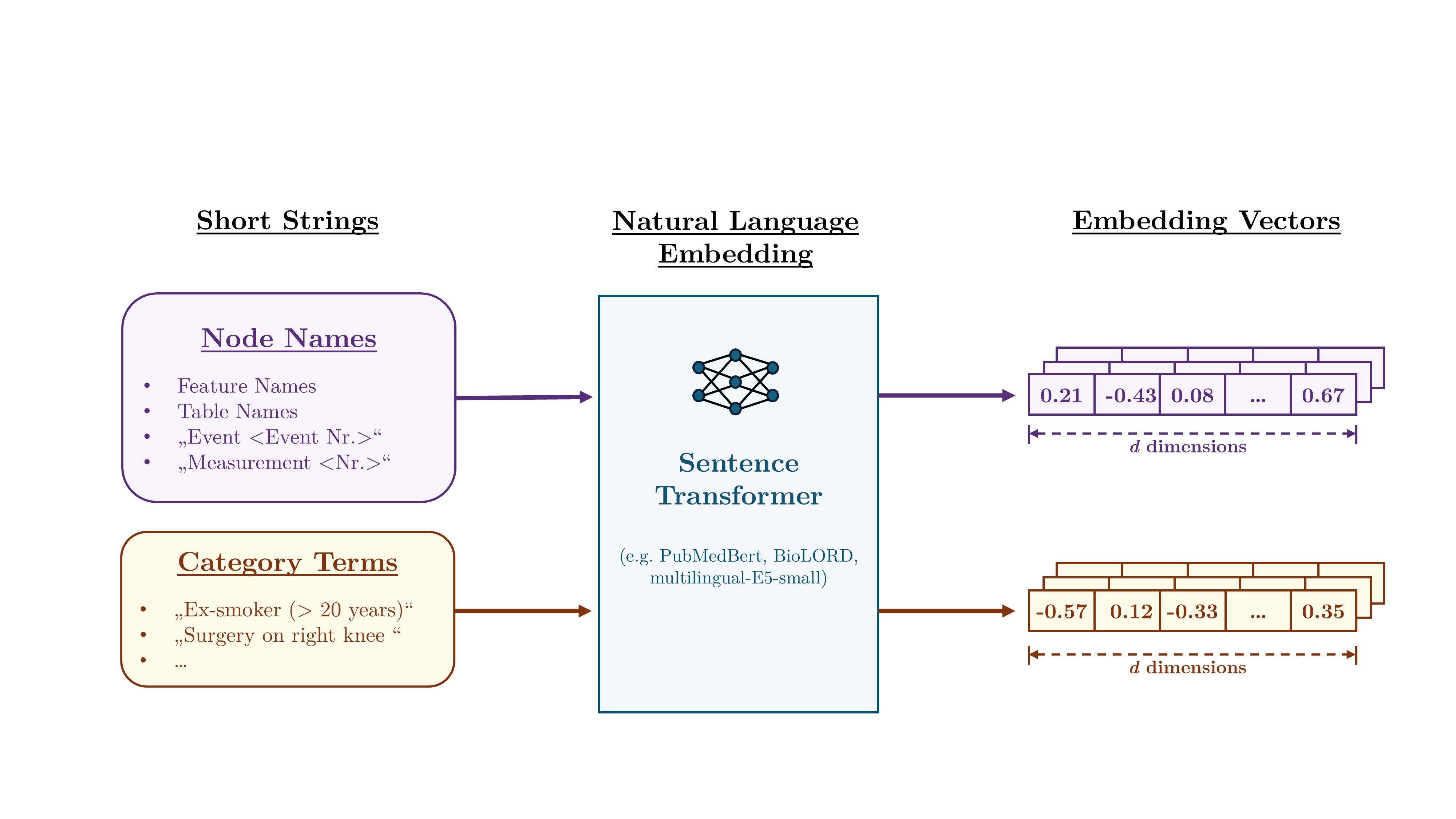}
    \caption{Embedding of node names and category terms as $d$-dimensional embedding vectors using various sentence transformers. The embedding dimension $d$ is determined by the sentence transformer used.}
    \label{fig:EmbeddingProcess}
\end{figure*}

To enable the direct application of GNNs on the PatTrees, we need node property vectors of uniform dimensionality across all nodes. As a basis for such vectors, we apply a Sentence Transformer on node names and category terms to retrieve fixed-dimensional embeddings, i.e. vectors of uniform length each representing the name of a node or one of the string values of a categorical features (c.f. figure \ref{fig:EmbeddingProcess}). Besides preparing the direct application of GNNs on the PatTrees, the embeddings provide the semantics of node names and category terms for eventual classification reasoning. Our ablation studies required different versions of embeddings to be retrieved. More details on these different versions can be found in section \ref{exp.emb}.

\subsection{PatTree Construction}\label{meth.kg_constr}
\begin{figure}
  \centering
    \includegraphics[width=\columnwidth]{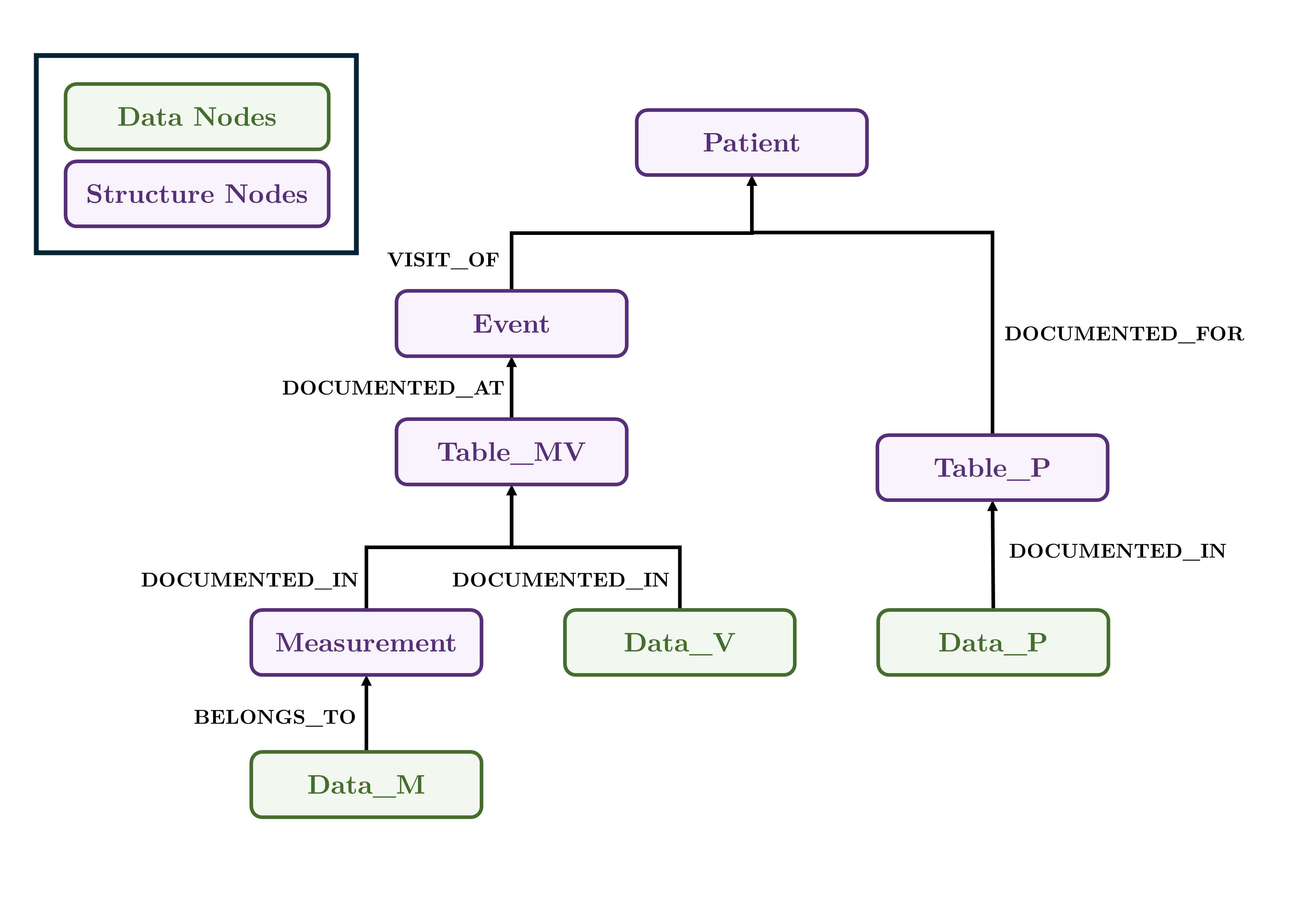}
    \caption{KG Schema behind PatTree construction with node types as indicated in Table \ref{tbl.node_names} and illustrative edge relationships as shown in Table \ref{tbl.edges}}\label{fig:KG_schema}
\end{figure}

For each patient, a PatTree is constructed from clinical data tables in three steps: (1) base graph creation, (2) adding data nodes, and (3) assigning node property vectors to each node. In this context, clinical data tables comprise raw data tables as well as tables with features extracted from other modalities such as images. The only assumption made is that the rows of each data table represents either a patient (P-centered table), a visit or event (V-centered table), or a measurement instance (M-centered table). The table type is automatically inferred from the data table based on the characteristics indicated in Table \ref{tbl.table_types}. Eventually, the constructed PatTree complies with the KG Schema visualized in Fig. \ref{fig:KG_schema} representing a directed acyclic graph with labeled edges. Here, edge directions and labels are structural annotations rather than components of the classification methodology. Edge directions only specify the direction of message passing in the subsequent graph classification step, and edge labels only encode the semantic structure of the graph for interpretability. They are not processed by the model, and edge directions have no function outside message passing. So, we can disregard them before message passing and results interpretation without loss of functionality. The pure data representation then reduces to a tree, an undirected, connected, acyclic graph \cite{valiente2002treesandgraphs}, which we refer to as the patient representation PatTree. In the following, references to root, leaf, parent, or child nodes adopt this tree-based interpretation: data nodes are leaf nodes, the central patient node is the root, and, for example, MEASUREMENT nodes are children of their corresponding TABLE nodes. The depth of a node is the minimal number of edges that need to be traversed until reaching the root node.

\subsubsection{Base Graph Creation}\label{meth.base_graph_creation}
For each patient, we firstly create a base graph containing most of the structure nodes: the central patient node, i.e. the root node, the event nodes, and the table nodes. Node names are indicated by Table \ref{tbl.node_names}. Each event node represents a date on which values were recorded for the patient in at least one of the tables. To also allow adding values for which the documentation date is unknown, we add an event node for "date unknown". For each event node, we create table nodes which represent the V- and M-centered data tables in which values were recorded at that event. Additionally, we create a single node for each P-centered data table containing values for the current patient. Eventually, we add edges to the base graph as indicated in Table \ref{tbl.edges}. Relation "VISIT\_OF" connects all created event nodes with the central patient node; relation "DOCUMENTED\_FOR" connects nodes representing P-centered tables with the central patient node; relation "DOCUMENTED\_AT" connects nodes representing M- and V-centered clinical tables with the node representing the respective event. Notably, edge directions and labels are included for illustrative and structural clarity only. They are not used by the downstream methodology except from edge directions indicating the direction of eventual message passing.

\subsubsection{Adding Data Nodes}
The second step of the PatTree construction process creates a data node for each value or extracted feature available. Based on the type of the table a data element stems from, we differentiate between patient-specific data, visit-specific data, and measurement-specific data which are loaded to the PatTrees in nodes of type Data\_P, Data\_V, and Data\_M, respectively. For each row read from an M-centered data table, we create a node of type Measurement, serving as the parent nodes for all data elements from that row. Eventually, the newly created nodes are integrated into the base graph via edges as indicated in Table \ref{tbl.edges}. Relation "BELONGS\_TO" connects nodes of type Data\_M with the respective measurement node; relation "DOCUMENTED\_IN" connects measurement nodes, Data\_V and Data\_P nodes with the respective table node from the base graph. As mentioned in section \ref{meth.base_graph_creation}, edge directions and labels are included for illustrative and structural clarity only.

\subsubsection{Assigning Node Property Vectors}\label{meth.node_property_vectors}
In the third and final step of PatTree construction, we assign node property vectors to each node in each PatTree. For this, we leverage outcomes of the data preprocessing outlined in \ref{meth.preprocessing}. The sentence embeddings described in \ref{meth.name_and_term_emb} are used to create initial node property vectors. For structure nodes, the node property vectors are initialized with the embedding vector of the node name. For data nodes, the node property vectors are a combination of the node name embeddings and the respective feature value or its embedding. Details are outlined in section \ref{exp.abl_stud}. The central patient node serves as the root node aggregating all information in the subsequent GAT-based classification outlined in section \ref{meth.KG_class}. Since the GAT applies sum aggregation, we assign the zero vector, i.e. the neutral element for addition, to the central patient node.

\begin{table}[ht]
\caption{Inference of data table type based on what the rows represent}\label{tbl.table_types}
\begin{tabular*}{\tblwidth}{@{}LLL@{}}
\toprule
Table Type & Rows represent... & Characteristics\\ 
\midrule
P-centered & Patients & \#patients equals \#rows\\
V-centered & Visits/ Events & \#events equals \#rows\\
M-centered & Measurements & \#rows/\#events $>$ 1\\
\bottomrule
\end{tabular*}
\end{table}

\begin{table}[ht]
\caption{Node types in PatTree and the node names assigned to nodes of the respective type. The text in angle brackets serves as a placeholder for the values of the respective node.}\label{tbl.node_names}
\begin{tabular*}{\tblwidth}{@{}LL@{}}
\toprule
Node Type & Node Name\\ 
\midrule
PATIENT & "Central Patient Node"\\
EVENT & "Event <Event Nr.>"\\
TABLE\_MV & "Table <Table Name>"\\
TABLE\_P & "Table <Table Name>"\\
MEASUREMENT & "Measured Instance <Measurement Nr.>"\\
DATA\_P & "<Feature Name>"\\
DATA\_V & "<Feature Name>"\\
DATA\_M & "<Feature Name>"\\
\bottomrule
\end{tabular*}
\end{table}

\begin{table}[ht]
\caption{Illustrative edge relationships in PatTree}\label{tbl.edges}
\begin{tabular*}{\tblwidth}{@{}LLL@{}}
\toprule
Head & Relationship & Tail \\ 
\midrule 
Event & VISIT\_OF & Patient\\
Table\_P & DOCUMENTED\_FOR & Patient\\
Table\_MV & DOCUMENTED\_AT & Event\\
Measurement & DOCUMENTED\_IN & Table\_MV\\
Data\_V & DOCUMENTED\_IN & Table\_MV\\
Data\_P & DOCUMENTED\_IN & Table\_P\\
Data\_M & BELONGS\_TO & Measurement\\
\bottomrule
\end{tabular*}
\end{table}

\subsection{PatTree Classification}\label{meth.KG_class}
The resulting PatTree can be mined by GNNs such as GATs directly, making it an easily usable multimodal patient representation. To demonstrate this, we interpret PatTree as a Feed Forward Network passing the node property vectors from leaf nodes through the graph towards the root node iteratively combining them with the node property vectors of other nodes within the same subtree. Finally, after four message passing and updating steps, each leaf node property vector reached the root node, i.e. the central patient node. This is a direct consequence of the PatTree construction process allowing a maximal depth of four. Therefore, we can interpret the node property vector assigned to the central patient node after four message passing and updating steps as the embedding vector representing the whole PatTree. Applying a final classification layer on this embedding vector yields the eventual classification.

\begin{figure*}
    \centering
    \includegraphics[width=\textwidth]{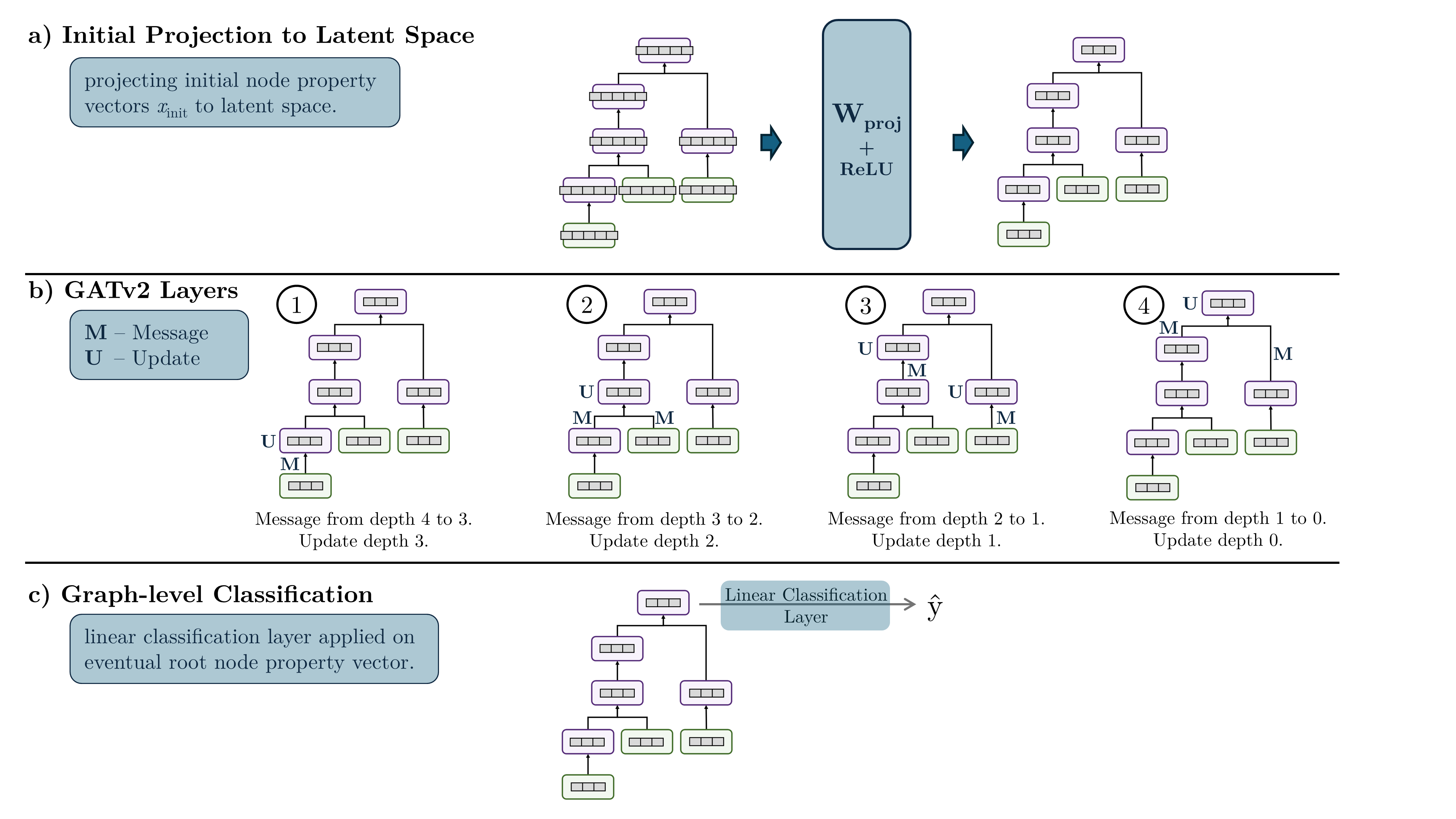}
    \caption{Three-stage GNN architecture applied to classify the PatTree patient representations as described in Section \ref{meth.KG_class}. a) Projection of initial node property vectors $x_{init}$ to hidden dimension $d_{hidden}$ (c.f. section \ref{meth.gnn_init}). b) GATv2-based message passing from leaf nodes to root in 4 GNN layers (c.f. sections \ref{meth.gnn_mssg}, \ref{meth.gnn_agg}, and \ref{meth.gnn_update}). c) Linear classification layer applied on final root node property vector for graph-level classification (c.f. section \ref{meth.gnn_classification}).}\label{fig:GNN_architecture}
\end{figure*}

For the message passing GNN, we chose a customized version of the GATv2 only updating the node property vectors at a currently active depth that moves from the parents of leaf nodes to the root node. The GNN firstly applies a linear layer projecting the initial vectors into a lower-dimensional latent space, after which four message passing GATv2 layers move the information carried by individual data features to the root node. Finally, the GNN applied a classification layer on the node property vector of the root node. Each of GATv2 layers implements a three-step message, aggregate and update scheme based on \textit{pytorch geometric}'s MessagePassing framework with attention-based sum aggregation over incoming neighbors in leaf-to-root direction. This direction of message passing corresponds to a source-to-target flow in PatTree's KG schema. After four layers, the vector of the root node integrates information from all nodes in PatTree. The full GNN architecture is depicted in figure \ref{fig:GNN_architecture}.

\subsubsection{Initial Projection to Latent Space}\label{meth.gnn_init}
The first layer is a linear transformation projecting the initial node vectors $\mathbf{x}_i^{\text{init}}$ of all nodes $i$ from dimension $d_{\text{in}}$ to $d_{\text{hidden}}$. $d_{\text{in}}$ equals the embedding dimension of the sentence transformer applied for embedding node names and category terms while $d_{\text{hidden}}$ is a GNN hyperparameter determining the dimensionality of the latent representations and the resulting full-graph embedding. The node property vector of node $i$ is now set to $\mathbf{x}_i^{0}$:
\begin{equation}
    \mathbf{x}_i^{0} = \text{ReLU}\left(\mathbf{W}_{\text{proj}} \, \mathbf{x}_i^{\text{init}}\right)
\end{equation}
where $\mathbf{W}_{\text{proj}}$ is a learnable linear transformation.

\subsubsection{Message Generation in GATv2 layers}\label{meth.gnn_mssg}
In total, there are 4 message passing GATv2 layers. In GATv2 layer $k$, with $k \in [1,4]$, $4-k$ is the currently active depth, i.e. the PatTree depth at which the node vectors are updated based on node property vectors received from depth $5-k$. For each edge $(j, i)$ with target node $i$ at active depth $4-k$, the current node property vector $\mathbf{x}^{(k-1)}_j$ of the source node $j$ is first linearly transformed into a message:

\begin{equation}
\mathbf{m}^{(k)}_j = \mathbf{W}^{(k)}_{\text{msg}}\, \mathbf{x}^{(k-1)}_j + b^k_{\text{msg}}
\end{equation}
where $\mathbf{W}^{(k)}_{\text{msg}}$ and $b^{(k)}_{\text{msg}}$ define a learnable linear transformation with bias term.

An attention coefficient is then computed between the transformed message $\mathbf{m}^{(k)}_j$ and the property vector $\mathbf{x}^{(k-1)}_i$ of the target node $i$, following the GATv2 formulation \cite{brody2021attentive}:

\begin{equation}
e^{(k)}_{ji} = \mathbf{a}^{(k)\top} \, \text{LeakyReLU}_{0.2}\left(\mathbf{W}^{(k)}_{\text{att}} \left[\mathbf{m}^{(k)}_j \, \| \, \mathbf{x}^{(k-1)}_i\right] + b^{(k)}_{\text{att}}\right)
\end{equation}
where $\mathbf{W}^{(k)}_{\text{att}}$ and $b^{(k)}_{\text{att}}$ define a learnable linear transformation with bias term, while $\mathbf{a}^{(k)}$ and $\|$ denote learnable edge scoring parameters and concatenation, respectively.

To restrict message passing to the child nodes of the currently active depth $4-k$, $e^{(k)}_{ji}$ is set to $-\infty$ for all edges whose target node $i$ is not at depth $4-k$, before applying a softmax over all incoming edges of target node $i$ to calculate the eventual attention score $\alpha^{(k)}_{ji}$:

\begin{equation}
\alpha^{(k)}_{ji} = \underset{j \, \in \, \mathcal{N}(i)}{\text{softmax}}\left(e^{(k)}_{ji}\right)
\end{equation}

For target nodes $i$ that receive no messages, the resulting $\alpha^{(k)}_{ji}$ is set to $0$ to avoid undefined attention scores. The final message sent from source node $j$ to target node $i$ is obtained by weighting the transformed message $\mathbf{m}^{(k)}_j$ with its corresponding attention score $\alpha^{(k)}_{ji}$:

\begin{equation}
\phi^{(k)}_{ji} = 
    \begin{cases}
        \alpha^{(k)}_{ji}\, \mathbf{m}^{(k)}_j, & \text{if}\ \text{depth}(i) = 4-k \\
        0, & \text{else}
    \end{cases}
\end{equation}

The final message $\phi^{(k)}_{ji}$ is then sent to target node $i$.
 
\subsubsection{Message Aggregation in GATv2 layers}\label{meth.gnn_agg}
The target node $i$ of an edge $(j, i)$ in the PatTree might receive several incoming messages if its in-degree is larger than 1. All these incoming messages are summed at the target node $i$ by aggregation function $\bigoplus$:
\begin{equation}
    \bigoplus\nolimits^{(k)}(i) = \sum_{j \, \in \, \text{children}(i)} \phi^{(k)}_{ji}
\end{equation}
 
\subsubsection{Node Vector Update in GATv2 layers}\label{meth.gnn_update}
After aggregation of incoming messages, the node property vector of target node $i$ at depth $4-k$ is updated by setting it to the output of update function $\gamma$:
\begin{equation}
    \gamma^{(k)}(i) = \text{ReLU}\left(\bigoplus\nolimits^{(k)}(i) + \mathbf{W}^{(k)}_{\text{upd}}\, \mathbf{x}^{(k-1)}_i\right)
\end{equation}
$\gamma^{(k)}(i)$ combines the aggregated incoming messages $\bigoplus^{(k)}(i)$ with a separately transformed version of the target node's own current property vector $\mathbf{x}^{(k-1)}_i$, using a learnable weight matrix $\mathbf{W}^{(k)}_{\text{upd}}$, before applying a joint $\text{ReLU}$ nonlinearity. This ensures that the current target node information is fused with the incoming messages from its child nodes, rather than being transformed independently.

Since only nodes at depth $4-k$ should be updated in layer $k$, the new node property vector $\mathbf{x}^{(k)}_i$ is obtained by applying $\gamma^{(k)}(i)$ selectively:
\begin{equation}
    \mathbf{x}^{(k)}_i =
    \begin{cases}
      \gamma^{(k)}(i), & \text{if}\ \text{depth}(i) = 4-k \\
      \mathbf{x}^{(k-1)}_i, & \text{else}
    \end{cases}
  \end{equation}
    
Thus, target nodes at the active depth $4-k$ receive the updated vector $\gamma^{(k)}(i)$, while all other nodes retain their previous property vector $\mathbf{x}^{(k-1)}_i$ unchanged.
 
\subsubsection{Eventual Graph-level Classification}\label{meth.gnn_classification}
 
Rather than pooling across all node embedding using a global readout function, we extract the final node property vector of the central patient node after four GNN layers as a full-graph embedding. This embedding is passed through a linear classification layer to produce logits over all classes:
\begin{equation}
    \hat{\mathbf{y}} = \mathbf{W}_{\text{cls}}\, \mathbf{x}_{\text{central\_patient\_node}}^{4} + b_{\text{cls}}
\end{equation}
Eventually, each sample is assigned the class with the highest logit value.

\section{Experimental Setup}\label{exp}
\subsection{Data}
In a first usability study, the PatTree methodolgy was tested in the use case of AD classification. We retrieved data from the Alzheimer's Disease Neuroimaging Initiative (ADNI), specifically the ADNI-1 cohort. ADNI is a large-scale, longitudinal multi-site study launched in 2003 with the primary goal of identifying biomarkers for the early detection and tracking of Alzheimer's disease progression \cite{mueller2005adni}. Data used in the preparation of this article were obtained from the ADNI database \footnote{adni.loni.usc.edu}.
The ADNI-1 cohort was selected for this work due to its high availability and completeness of sMRI data. The dataset employed comprises a total of 763 patients, for whom both clinical assessments and T1-weighted structural MRI scans were available. Clinical data included demographic information, family history, examination results, lab test results, vital signs, clinical history, cognitive evaluation scores, diagnostic labels, and relevant neuropsychological assessments as provided by the ADNI data repository. We exported all data tables available for the ADNI1 cohort. Initial data exploration revealed a class imbalance with class MCI being over-represented (AD: 184, MCI: 368, CN: 211) as well as a gender bias manifesting in an over-representation of male patients (Male: 447, Female: 316).
The age of the patients spreads between 55 and 90 years with a mean of 75.2 years and a standard deviation of 6.61 years.

\subsection{Experimental Design}\label{exp.design}
Our main goal is to evaluate the usability of the PatTree representation in the ADNI classification setting, rather than developing an optimal ADNI classifier. Through data preprocessing (\ref{exp.data_prepro}) and PatTree construction (\ref{exp.KG_constr}), we retrieved one PatTree for each patient. The PatTrees were classified by use of a customized GATv2 model as defined in section \ref{meth.KG_class}. We considered two different classification settings: binary (AD vs. CN) and multiclass (AD vs. MCI vs CN). The complete cohort of 763 patients was split into a training, validation and test set with 492, 121, and 150 patients, respectively. In the two class setting, the sizes of training, validation and test set are reduced to 251, 68, and 76, respectively. The classification results were compared to a baseline transformer model classifying the patients based on the preprocessed data tables directly, stopping before PatTree creation. Additionally, we systematically  investigated the impact of individual components by conducting ablation studies on the following ablation targets:
\begin{enumerate}
    \item Sentence Transformer: sentence transformer applied to retrieve embeddings of node names and category terms,
    \item Data Node Vectors: the strategy for construction of initial node property vectors for data nodes, i.e., how the node name embedding is combined with the data value,
    \item Structure Node Vectors: the choice of node property vectors for structure nodes, i.e. name embeddings versus zero vectors,
    \item Image Features Included: the number and type of image features included.
\end{enumerate}
All ablation studies revolve around choices in PatTree construction instead of GAT parameters: This focus aligns with our main goal to investigate the usability of the PatTree representation.
More details on the realization of the ablation studies can be found in section \ref{exp.abl_stud}.
Exhaustively testing all options and combinations of options across the four ablation studies in two classification settings (binary and three-class), we ended up training and testing 144 PatTree-based GNN models. Given the resulting scale, we restricted our analysis to the respective peak performance overall and in individual ablation studies. Since our primary aim is examining whether the PatTree representation can be meaningfully mined at all, it suffices to show that at least one model configuration is able to do so. Identifying optimal configurations would require monitoring the full distribution of outcomes across all configurations which remains as future work.

\subsection{Data Preprocessing}\label{exp.data_prepro}
Prior to model development, the raw ADNI-1 clinical and imaging data underwent a systematic preprocessing pipeline to ensure data quality and suitability for downstream analysis.
Several fields serving purely administrative or tracking purposes were excluded from the dataset, specifically: ID, RECNO, SITEID, VISCODE, VISCODE2, RID, USERDATE, and USERDATE2. These fields carry no clinically or diagnostically relevant information and were therefore removed to reduce noise.
Data fields flagged with a status of Archived, Redacted, or Deprecated in the ADNI data dictionary were systematically excluded as well.
To facilitate interpretability and meaningful text embeddings, each downloaded ADNI table was assigned a human-readable name. These names were extracted from the CRFNAME column of the ADNI data dictionary, which documents the natural language table name associated with each data table.
Columns containing no observations i.e., entirely missing across all subjects were dropped from the dataset.
Features for which no textual description was available in the ADNI data dictionary were excluded. This criterion ensures that all retained features are semantically interpretable, which is a prerequisite for the creation of the text embedding-based node property vectors employed in PatTree.
Continuous features, including tabular clinical variables and features extracted from structural MRI scans, were normalized to a common scale using $z$-score normalization. Categorical variables were replaced by the natural language descriptions of their respective category labels as documented in the ADNI data dictionary, rather than being encoded as arbitrary numeric codes. This representation renders categorical information semantically coherent for later text-based processing.
The clinical data were filtered to retain only those subjects for whom a corresponding structural MRI scan was available. This step ensured consistency between the clinical and imaging modalities, yielding a fully paired dataset of 763 subjects.
The diagnosis table was excluded from the feature input space to prevent target leakage. Class labels (CN, MCI, AD) were extracted from this table separately and used exclusively as the prediction targets for subsequent classification experiments.

\subsubsection{Image Feature Extraction}\label{exp.img_feat_extr}
Motivated by prior work in AD classification \cite{henschel2020fastsurfer, bloch2021comparison, de2024pipnet3d} and following the methodology for image feature extraction outlined in \ref{meth.img_feature_extr}, we apply an unsupervised and a supervised approach for image feature extraction from MRI scans. The first, unsupervised approach applied the deep-learning tool FastSurfer for segmenting the full brain into 95 ROIs listed by the DKTatlas and, subsequently, calculating the volumes of these brain regions \cite{henschel2020fastsurfer}. Afterwards, we merged brain regions on lobe-level and passed the lobe-level images to the second, supervised image feature extraction approach. This second approach was realized with the PIPNet3D architecture for extraction of prototype-based image features for each ROI \cite{de2024pipnet3d}. For this, a 3D-ResNet-18 pretrained on the Kinetics-400 dataset was combined with a subsequent max-pooling layer and a linear layer performing calculation of prototype similarity scores and AD classification, respectively, for each of the 95 ROIs. In a pretraining step, the parameters of each ResNet were finetuned to the ADNI training data while the linear layer was frozen. In a subsequent main training step, the parameters of all layers were optimized using a weighted loss. All training steps were performed using phase- and ROI-specific hyperparameters determined by hyperparameter optimization and recommendations by Nauta et al. \cite{nauta2023pip}. The training and inference processes of the 95 PipNet3D models ran on a dual socket server with two Intel(R) Xeon(R) Silver CPUs (2.30GHz, 503GiB RAM) as well as a NVIDIA A16 GPU with 64 GiB VRAM.

The lobe-level ROI volumes and prototype similarity scores of segmented MRI scans can be interpreted as continuous features representing the MRI scans. As such features, they facilitate the inclusion of information from medical images as numerical values in the PatTree. Due to computational complexity, we only extracted the image features from ADNI's baseline assessment MRI scans.

\subsubsection{Embedding of Node Names and Category Terms}\label{exp.emb}
As outline in section \ref{meth.name_and_term_emb}, we applied pretrained sentence transformers to create embeddings of node names and category terms. To prepare the ablation study on different node property vectors for data nodes, we additionally created long node names. These long node names are a concatenation of the node name and the string representation of the respective feature value in the data node. For instance, the long node name of a data node with node name "Body height" and value "180cm" would be "Body height 180cm". We embedded all node names, long node names and category terms extracted from the data tables. To enable investigating the impact of the choice of the sentence transformer, we applied three different, pretrained models. In particular, we chose two biomedical sentence transformers with embedding dimension 768: PubMedBert\footnote{neuml/pubmedbert-base-embeddings} and BioLORD\footnote{FremyCompany/BioLORD-2023} \cite{gu2021domain, remy-etal-2023-biolord}. As a control, we additionally applied a domain-agnostic sentence transformer with embedding dimension 384: Multilingual-E5-small\footnote{intfloat/multilingual-e5-small} ("Small" in the following) \cite{wang2024multilingual}. All models were retrieved from HuggingFace through library \textit{sentence\_transformers} and applied without further modification. The models were deployed on two Intel(R) Xeon(R) Silver CPUs (2.30GHz, 503GiB RAM) and a NVIDIA A16 GPU (64 GiB VRAM).

\subsection{PatTree Construction}\label{exp.KG_constr}
\subsubsection{PatTree Construction for the ADNI dataset}
\begin{figure}
    \centering
    \includegraphics[width=\linewidth]{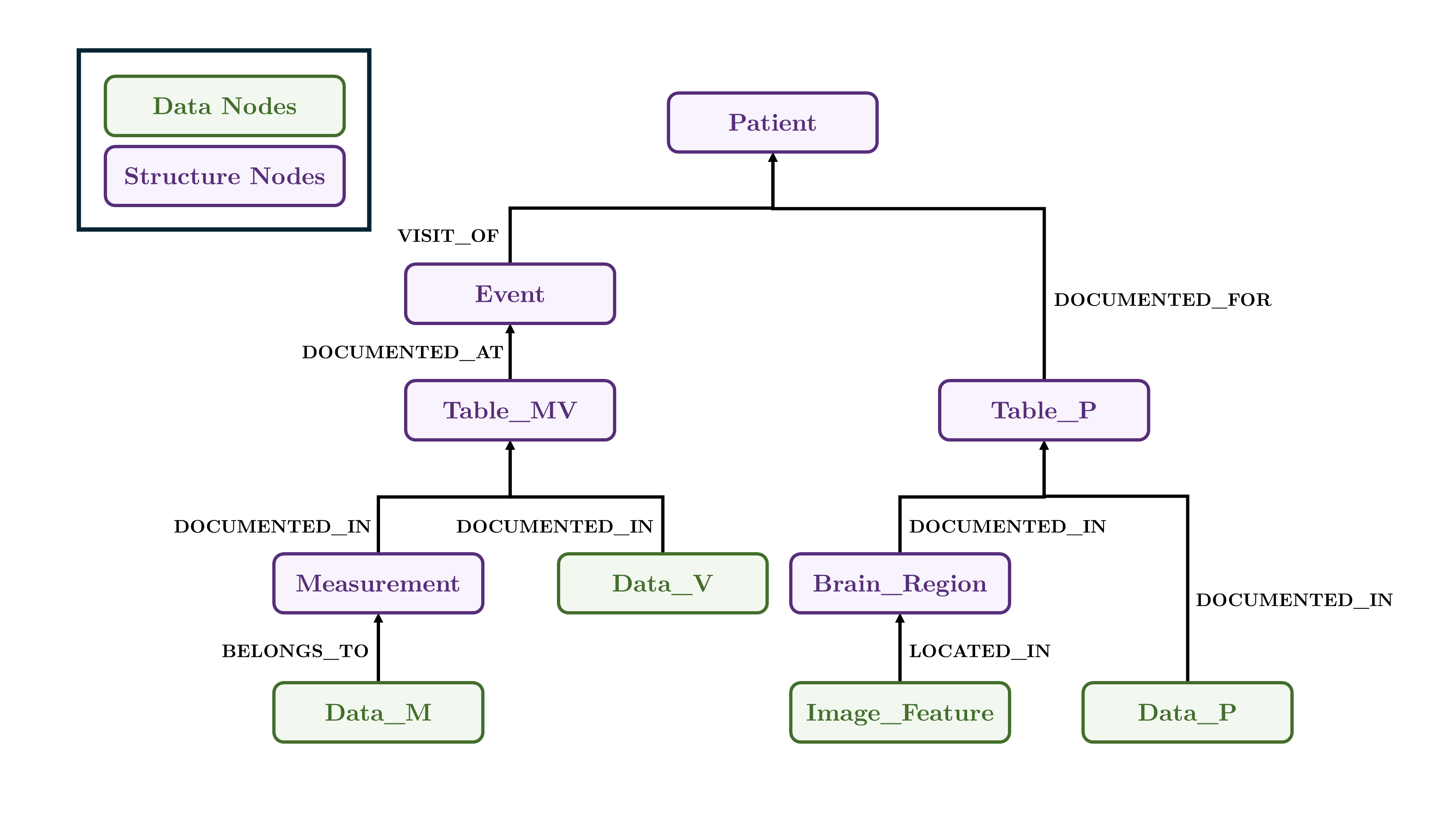}
    \caption{KG Schema behind PatTree construction from Figure \ref{fig:KG_schema} adapted for our Image Feature Extraction process in the ADNI Use Case.}
    \label{fig:KG_schema_ADNI}
\end{figure}

We created a PatTree for each patient using python library \textit{networkX} v3.5 mainly following the methodology outlined in \ref{meth.kg_constr}. The only adaption was creating two new node types. Node type Brain\_Region accommodates the lobe-level ROIs for which image features were calculated in \ref{exp.img_feat_extr}; nodes of type Image\_Feature represent calculated ROI volumes or prototype similarities. Relation "DOCUMENTED\_IN" connects nodes of type Brain Region to a node of type Table\_P representing the table of extracted image features; relation "LOCATED\_IN" connects the Image Feature nodes with the Brain\_Region node representing the ROI for which they were calculated. These adaptions were made because we only extracted the image features from ADNI's baseline assessment MRI scans meaning that we only had a single set of image features per patient making the image features patient-centered but still groupable by ROI. The resulting KG schema is depicted in figure \ref{fig:KG_schema_ADNI}. The PatTree construction was performed on an Intel Core Ultra 7 155U (1.70 GHz, 32GB RAM).

\begin{figure*}
    \centering
    \includegraphics[width=0.9\textwidth]{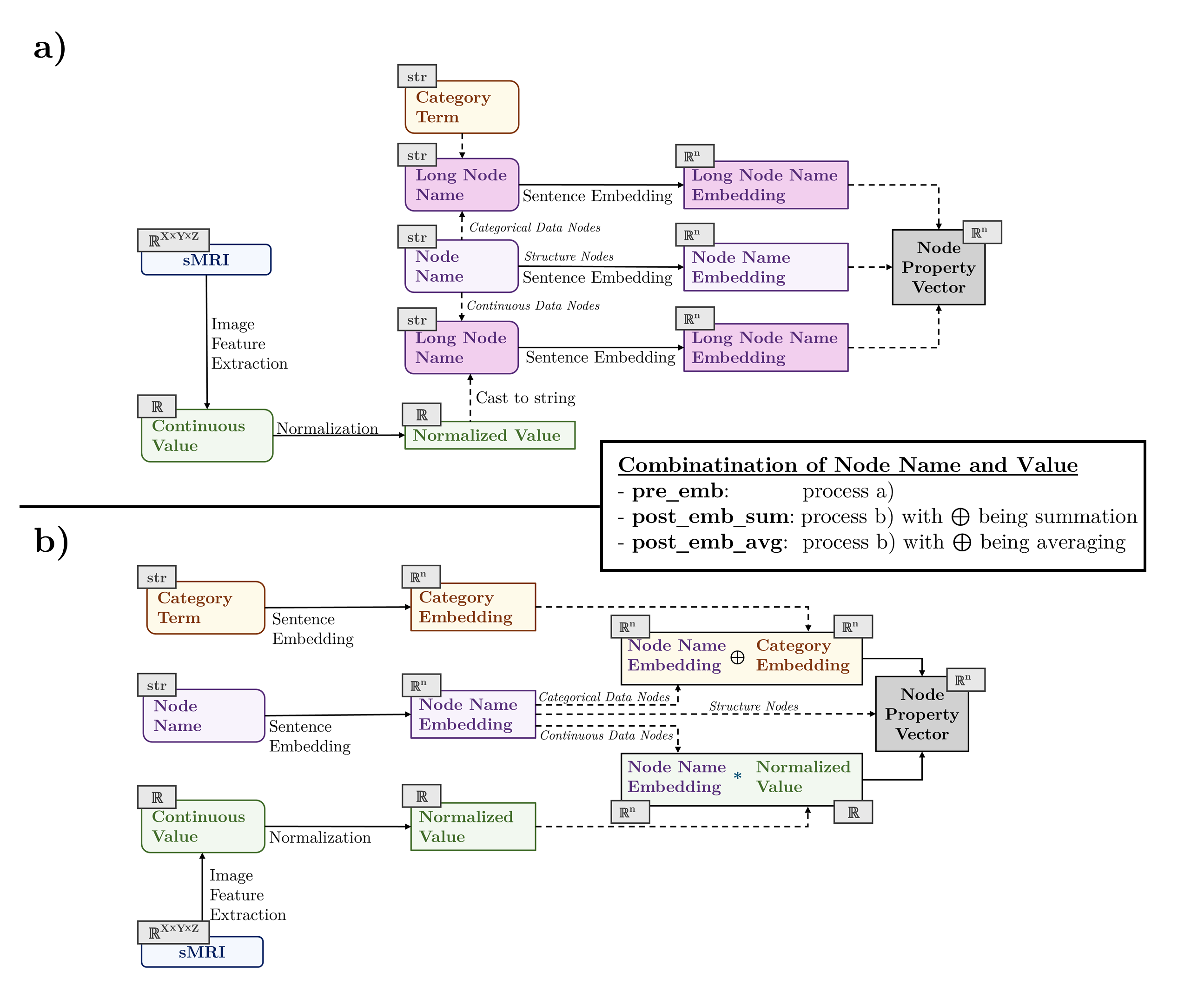}
    \caption{Processes applied for creating initial node property vectors from heterogeneous data features and metadata as described in Section \ref{exp.abl_stud}. a) Process for creation of initial node property vectors using strategy "pre\_emb" for combination of node names and data values. b) Process for creation of initial node property vectors using "post\_emb" strategies for combination of node names and data values.}\label{fig:Node_vector_creation}
\end{figure*}

\subsubsection{Preparing Ablation Studies}\label{exp.abl_stud}
Besides creating the basic PatTree, the construction process prepared the ablation studies outlined in \ref{exp.design}. Firstly, each node was assigned nine initial node property vectors. This is a result of applying three different sentence transformers (PubMedBert, BioLORD, Small) and, subsequently, three different strategies for combining node name embeddings with data node values (ablation studies 1 and 2). As outlined in section \ref{meth.node_property_vectors}, the node property vectors of structure nodes are always initialized with the pure embedding of the node name. The node property vector of the central patient node is always initialized with the zero vector.

For data nodes, the node property vectors are a combination of the node name embeddings and the respective feature value or its embedding. The applied combination strategies are depicted in figure \ref{fig:Node_vector_creation}. The first strategy, referred to as "pre\_emb", fuses node names and data values prior to string embedding into a single string which is then embedded. This means that we assign the embeddings of the long node names described in section \ref{exp.emb} as the initial feature vectors. The second and third strategy fuse node name and data value after string embedding with a slightly different approach for categorical and continuous features. The image features resulting from the process described in \ref{meth.img_feature_extr} are used as continuous values representing the MRI scans. For all data nodes carrying continuous values, including those resulting from the image feature extraction described in \ref{meth.img_feature_extr}, the node property vector is built by multiplying their $z$-score normalized value with the embedding vector representing their node name. For data nodes representing categorical values, the node property vector is set to the combination of the embedding vectors representing their node name and their category term. Using summation to combine the embedding vectors, results in strategy two referred to as "post\_emb\_sum"; using averaging of embedding vectors results in strategy three referred to as "post\_emb\_avg".

Eventually, we implemented a customizable PatTree loading process enabling the user to choose the node property vectors as well as which image features to include. The choice of node property vectors included the procedure of how the embeddings of data node names were combined with data values as well as the choice to replace node property vectors of structure nodes by zero vectors.

\subsection{PatTree Classification}\label{exp.KGclassification}
For classification of the created PatTrees we trained several GNNs based on the customized GATv2 architecture described in section \ref{meth.KG_class}, i.e. passing information through the graph iteratively, aggregating all information in the central patient node, and, eventually, classifying the graph based on the final node property vector of the central patient node. The GNNs were implemented using python library \textit{pytorch geometric} v2.7.0.

To enable the ablation studies outlined in section \ref{exp.design}, we consider different PatTree versions. The versions differ in (1) the Sentence Transformer applied for initial node property vector creation (PubMedBert, BioLORD, Small), (2) the strategy for combining node names and data values for initial node property vectors in data nodes (pre\_emb, post\_emb\_sum, post\_emb\_avg), (3) the choice of initial node property vectors in structure nodes (node name embeddings, zero vector), and (4) the type and number of image features included (none, ROI volumes, prototype similarities, all). We trained one customized GATv2 for each combination of the four PatTree ablation targets, resulting in a total of 72 GNNs per classification setting. This whole procedure was run for two different classification settings: binary (AD vs. CN) and multiclass (AD vs. MCI vs CN). All GNNs were trained with the same set of hyperparameters: 50 epochs, batch size of 16 PatTrees, learning rate of 0.01, and a hidden layer size, i.e. $d_{hidden}$, of 32. Parameters were retrieved using an Adam optimizer and we applied a weighted cross-entropy loss to account for the observed class imbalance. Training and inference were performed on two Intel(R) Xeon(R) Silver CPUs (2.30GHz, 503GiB RAM) and a NVIDIA A16 GPU (64 GiB VRAM).

To evaluate model performance, we employed two complementary metrics: balanced accuracy (bal. ACC) and F$_1$ score. On a scale between 0 and 1, these metrics evaluate the weighted per-class effectiveness of a classifier and the relations between the data’s positive labels and those given by a classifier based on a per-class average, respectively \cite{sokolova2009systematic}. In the three-class setting, the F$_1$ score was calculated by macro-averaging class-specific F$_1$ scores (one vs. the rest). All trained models were assessed on both the training dataset, to monitor learning and detect overfitting, and the held-out test set, which was not included in any training or model selection process. This evaluation on the held-out test test provides an unbiased estimate of generalization performance.

\subsection{Baseline Classification}\label{exp.baseline}
To investigate the contribution of the PatTree representation on the classification performance, we trained a second set of attention-based models on the preprocessed data tables directly. These models serve as the baseline that we can compare the PatTree GNN performance to. Following the Feature Tokenizer + Transformer (FT-Transformer) approach by Gorishniy et al. \cite{gorishniy2021revisiting}, we trained one Transformer-based model per preprocessed data table classifying patients in the table. These models were implemented using python library \textit{torch} v2.6.0.

\subsubsection{Additional Preprocessing of Data Tables} 
To make the table values purely numeric, category terms were replaced by their embeddings retrieved in \ref{exp.emb}. This means that we retrieved three different versions of each table differing in the applied sentence transformer generating the category term embeddings. To obtain a single classification per patient with the FT-Transformer approach, we need one row per patient in each table. To this end, we aggregated table rows for each patient across their visits with mean pooling, i.e. averaging continuous values and category term embeddings across visits for V- and M-centered tables prior to classification. Empty rows were dropped and individually missing values were replaced by zero or the zero vector for continuous and categorical features, respectively. Replacing missing values in a $z$-score normalized column by 0 corresponds to mean imputation.

\subsubsection{Baseline Model}
Following the FT-Transformer approach \cite{gorishniy2021revisiting} but adapting it to the GATv2-based models applied on the PatTrees, we implemented a two-layer transformer for each data table with a single attention head. The individual transformers output class probabilities for each patient, i.e. each row resulting from mean pooling original rows for each patient. The classification is based on the information of the table they were trained on. Final, table-agnostic classifications were retrieved by averaging per-patient class probabilities across table-specific transformer outputs. All table-specific transformers were trained with the same set of hyperparameters: 20 epochs, batch size of 8, learning rate of 0.001, a hidden layer size of 64, and a dropout of 0.1. Just as in PatTree classification, parameters were retrieved using an Adam optimizer and we applied a weighted cross-entropy loss to account for the observed class imbalance. Training and inference were performed on a NVIDIA RTX A4000 GPU (16 GB VRAM) with an Intel Core i9-10920X CPU (3.50 GHz, 62 GB RAM).

\section{Experiment Results}\label{exp_res}
\subsection{Runtime}
\subsubsection{Graph Creation Process}\label{graph_creation_res}

The first step in graph creation was image feature extraction. The runtimes of this initial step are shown in Table \ref{tbl.img_feat_extr_runtime}. We observed a significantly longer runtime for the extraction of prototype similarities than for the calculation of ROI volumes, since the former required training the PIPNet3D model while the FastSurfer model could be used out-of-the-box; conversely, inference was slower for ROI volumes, though not enough to offset the training time invested in the prototype similarity approach.

As shown in Table \ref{tbl.preprocessing_runtime}, the preprocessing runtime was dominated by the processing and lobe-wise grouping of ROI volumes, with all other preprocessing steps completing in three seconds or less. The graph creation runtime depicted by Table \ref{tbl.graph_creation_runtime}) was comparatively balanced across its three stages base graph creation, addition of clinical data nodes, and addition of image feature nodes. However, adding clinical data nodes took slightly longer than the other steps, reflecting the larger number of nodes involved.

Comparing the different strategies for the creation of data node property vectors (Table \ref{tbl.embedding_runtime}, Table \ref{tbl.node_vectors_runtime}), the pre\_emb strategy was markedly the slowest during the embedding step due to the length of the node names being embedded. In node vector creation the pre\_emb strategy was faster than the post\_emb strategies since no combination of embeddings was required. However, this saving did not compensate for the additional embedding time. The two post\_emb startegies post\_emb\_sum and post\_emb\_avg, by design, showed no runtime differences during embedding, and no significant difference during initial node vector creation.

\begin{table}[ht]
\caption{Runtimes for training and inference of image feature extraction models. The training runtime is given in hours (h); the inference runtime is given in seconds (s).}\label{tbl.img_feat_extr_runtime}
\begin{tabular*}{\tblwidth}{@{}LCR@{}}
\toprule
Features & Training [h] & Inference [s] \\ 
\midrule 
ROI Volumes & 0 (pretrained) & 153.00 \\
Prototype Similarities & 78 & 0.63\\
\bottomrule
\end{tabular*}
\end{table}

\begin{table}[ht]
\caption{Mean training runtimes of steps in data preprocessing given in seconds (s)}\label{tbl.preprocessing_runtime}
\begin{tabular*}{\tblwidth}{@{}LR@{}}
\toprule
Step & Runtime [s] \\ 
\midrule 
Preprocess data dictionary & 0.238 \\
Retrieve class labels & 0.253\\
Preprocess clinical data & 3.276\\
Preprocess ROI volumes & 83.821\\
Preprocess prototype similarities & 0.317\\
Infer data table types & 1.261\\
\textbf{Total} & 89.166\\
\bottomrule
\end{tabular*}
\end{table}

\begin{table}[ht]
\caption{Mean training runtimes of steps in graph creation given in seconds (s)}\label{tbl.graph_creation_runtime}
\begin{tabular*}{\tblwidth}{@{}LR@{}}
\toprule
Step & Runtime [s] \\ 
\midrule 
Base Graph Creation & 22.736 \\
Adding Clinical Data Nodes & 65.629\\
Adding Image Data Nodes & 34.327\\
\textbf{Total} & 122.692\\
\bottomrule
\end{tabular*}
\end{table}

\begin{table}[ht]
\caption{Runtimes in seconds (s) of different pre-trained Sentence Transformers for embedding node names, long node names, and category terms}\label{tbl.embedding_runtime}
\begin{tabular*}{\tblwidth}{@{}LRRR@{}}
\toprule
Embedding Input & PubMedBert [s] & BioLORD [s] & Small [s]\\ 
\midrule 
Node Names & 7.010 & 7.538 & 5.489\\
Long Node Names & 1011.445 & 1078.514 & 896.920\\
Category Terms & 106.730 & 111.379 & 93.006\\
\textbf{Total} & 1125.185 & 1197.431 & 995.415\\
\bottomrule
\end{tabular*}
\end{table}

\begin{table}[ht]
\caption{Runtimes in seconds (s) for calculating initial node property vectors on embeddings from different Sentence Transformers and for different strategies for combining embeddings of node names and data values for data nodes}\label{tbl.node_vectors_runtime}
\begin{tabular*}{\tblwidth}{@{}LRRR@{}}
\toprule
Data Node Vectors & PubMedBert [s] & BioLORD [s] & Small [s]\\ 
\midrule
Pre\_emb & 916.645 & 998.668 & 513.847\\
Post\_emb\_sum & 1264.071 & 1303.410 & 715.507\\
Post\_emb\_avg & 1261.876 & 1297.270 & 692.177\\
\bottomrule
\end{tabular*}
\end{table}

\subsubsection{Training of Baseline Transformer}\label{res.baseline_training}

For each sentence transformer applied for category term embedding, we trained a baseline transformer on each data table classifying the patients. A fused classification was retrieved by mean aggregation across all data table-specific classifications (c.f. section \ref{exp.baseline}). The number of target classes and the number of tables fused were the main influential factors on the training runtime for these baseline transformers. As shown in Table \ref{tbl.training_runtime_baseline}, individual-table models trained in a mean of 2.9 seconds and a standard deviation of 1.0 in the binary setting. In the three-class setting, the mean training runtime was 5.6 seconds with a standard deviation of 1.9 seconds, reflecting the increased complexity of the multiclass task. Relative to their means, individual-table runtimes showed considerable variability with a standard deviation of roughly 34\% of the mean in both settings. This variability is likely driven by differences in table size such as the number of rows and columns. Fusing predictions across all tables naturally required proportionally longer cumulative training time, with mean runtimes of 82.3 seconds and a standard deviation of 2.6 seconds in the binary setting. In the three-class setting, the mean training runtime was 155.9 seconds with a standard deviation of 5.3 seconds. So, when cumulating training times of table-specific transformers, the relative variability reduced substantially to around 3\% of the mean as fluctuations in training time across individual tables tended to average out once aggregated.

\begin{table}[ht]
\caption{Mean training runtimes of baseline transformers in seconds (s) across all training runs on individual tables and the cumulative training runtimes for fused models. The standard deviation is given in parentheses.}\label{tbl.training_runtime_baseline}
\begin{tabular*}{\tblwidth}{@{}LLR@{}}
\toprule
Classification & Scope & Training time [s] \\ 
\midrule 
\multirow{2}{*}{Binary} & Table-specific & 2.9 (1.0)\\
& Fused classification & 82.3 (2.6)\\
\midrule
\multirow{2}{*}{Three-class} & Table-specific & 5.6 (1.9)\\
& Fused classification & 155.9 (5.3)\\
\bottomrule
\end{tabular*}
\end{table}

\subsubsection{Training of PatTree Models}\label{res.gnn_training}
Just as for the baseline transformer, the training runtime of the PatTree-based GNNs was determined primarily by the number of target classes, remaining stable at roughly six hours for the binary setting and roughly twelve hours for the three-class setting across all ablation studies. In the two-class setting, the 72 training runs averaged 21333 seconds (5 hours, 55 minutes, 33 seconds), with a standard deviation of 264 seconds (4 minutes, 24 seconds); as Table \ref{tbl.training_runtime_binary} shows, this runtime remained stable across ablation studies with only minimal deviation. In the three-class setting, the 72 runs averaged 43148 seconds (11 hours, 59 minutes, 8 seconds), with a standard deviation of 437 seconds (7 minutes, 17 seconds), again showing stable runtimes across ablation studies (Table \ref{tbl.training_runtime_threeClass}).

\begin{table}[ht]
\caption{Mean training runtimes of the PatTree GNNs in the binary classification setting given in seconds (s) across all training runs with the respectively specified setting. The standard deviation is given in parentheses.}\label{tbl.training_runtime_binary}
\begin{tabular*}{\tblwidth}{@{}LLR@{}}
\toprule
Ablation Target & Setting & Training time [s] \\ 
\midrule 
\multirow{3}{*}{Sentence Transformer} & PubMedBert & 21253 (189)\\
& BioLORD & 21294 (212)\\
& Small & 21453 (334)\\
\hline 
\multirow{3}{*}{Data Node Vector} & Pre\_emb & 21268 (191)\\
& Post\_emb\_sum & 21273 (215)\\
& Post\_emb\_avg & 21459 (328)\\
\hline 
\multirow{2}{*}{Structure Node Vector} & Zero & 21382 (279)\\
& Embedding & 21285 (243)\\
\hline 
\multirow{4}{*}{Image Features} & None & 21363 (278)\\
& ROI volumes & 21297 (176)\\
& Prototype Sim. & 21102 (211)\\
& All & 21572 (140)\\
\bottomrule
\end{tabular*}
\end{table}

\begin{table}[ht]
\caption{Mean training runtimes of the PatTree GNNs in the three-class classification setting given in seconds (s) across all training runs with the respectively specified setting. The standard deviation is given in parentheses.}\label{tbl.training_runtime_threeClass}
\begin{tabular*}{\tblwidth}{@{}LLR@{}}
\toprule
Ablation Target & Setting & Training time [s] \\ 
\midrule 
\multirow{3}{*}{Sentence Transformer} & PubMedBert & 42971 (267)\\
& BioLORD & 43405 (431)\\
& Small & 43067 (473)\\
\hline 
\multirow{3}{*}{Data Node Vector} & Pre\_emb & 43204 (428)\\
& Post\_emb\_sum & 43145 (404)\\
& Post\_emb\_avg & 43094 (486)\\
\hline 
\multirow{2}{*}{Structure Node Vector} & Zero & 43256 (450)\\
& Embedding & 43040 (401)\\
\hline 
\multirow{4}{*}{Image Features} & None & 43330 (505)\\
& ROI volumes & 43009 (301)\\
& Prototype Sim. & 42930 (469)\\
& All & 43322 (307)\\
\bottomrule
\end{tabular*}
\end{table}

\subsection{Classification Performance}
\subsubsection{Baseline Performance}\label{baselinePerf}
The baseline approach applied a FT-transformer-inspired model to each preprocessed data table individually, producing one classification per table and patient, which was then combined via averaging per-patient class probabilities across table-specific transformer outputs (c.f. section \ref{exp.baseline}). Table \ref{tbl.baselinePerf_individual} reports the best individual-table performance, and Table \ref{tbl.baselinePerf_fused} the best performance after combining table-specific predictions. Notably, in the three-class setting no single table matched the predictive performance of the fused prediction on the held-out test set.

In the binary setting, the tables FNIHBC and MMSE lead to perfect classification performance (bal. ACC and F$_1$ score of 1.0 on the test set) regardless of the embedding model used; the model trained on the CDR table achieved the same perfect performance using the PubMedBert or Small embeddings, and model trained on the NEUROPATH table reached it using the BioLORD or Small embeddings. This suggests a high predictive power of these listed data tables, consistent with recent related findings \cite{diagnostics16121755}. As a result, the fused classification exhibited perfect classification performance as well, with the single exception of the transformer using the BioLORD embeddings, whose best fused result reached a bal. ACC of 98.5\% and an F$_1$ score of 0.987. In the three-class setting, the strongest individual-table result on the test set came from the CDR table using the Small embeddings (bal. ACC 86.5\%, F$_1$ score 0.844), while the best fused performance reached a bal. ACC of 88.7\% and an F$_1$ score of 0.868. It was achieved without considering any image features and using BioLORD category-term embeddings.

To monitor potential overfitting, we additionally examined classification performance on the training set, which was, as expected, generally higher than or equal to performance on the test set. In the binary setting, the tables FNIHBC and our extracted prototype similarities achieved perfect classification on the training data across all embedding models, and the model on the CDR table reached perfect training performance using the BioLORD embeddings. In the three-class setting, the prototype similarities table stood out as the only table achieving consistently strong training performance, with bal. ACC ranging from 99.4\% (Small) to 100\% (PubMedBert) and F$_1$ scores from 0.996 (Small) to 1.000 (PubMedBert). However, this performance was not reproduced on the test set, suggesting that the prototype similarities table is particularly prone to overfitting in the three-class setting which might be a result from the supervised nature of the feature extraction process obtaining the prototype similarities.

Regarding embedding model choice (Table \ref{tbl.baselineSentenceTransf}), all fused models in the binary setting that used the embeddings of sentence transformers PubMedBert and Small reached perfect performance, while the BioLORD embeddings lead to slightly decreased performance (98.5\% bal. ACC, F$_1$ score 0.987). However, this overall strong binary performance is largely attributable to the high predictive power of the tables listed above. In the more challenging three-class setting, the BioLORD embeddings performed best on the test set (88.7\% bal. ACC, F$_1$ score 0.868).

In terms of image modalities (Table \ref{tbl.baselineImgMod}), the fused models performed best across all settings when no image features were included. Adding classification results on any of the image features to the fused model consistently reduced baseline performance although related work generally suggests that additional modalities should improve classification performance, as outlined in section \ref{rl.multmod}. Potential reasons for the reduced performance when including image features are the choice of image feature extraction procedure itself or to how these features were integrated into the baseline classifier, rather than to a genuine lack of value in imaging data.

In summary, the tables FNIHBC and MMSE are highly predictive in the binary classification setting, as are the tables NEUROPATH and CDR and, at least on the training set, the prototype similarities table. The perfect classification results on these tables in the baseline limit the interpretability of results in the binary setting. In the three-class classification setting, by contrast, no table achieved perfect prediction on the test set, suggesting these results are more meaningful. The best overall baseline performance was a bal. ACC of 88.7\% and an F$_1$ score of 0.868, achieved in the three-class setting.

\begin{table}[ht]
\caption{Peak classification performance of baseline transformer models across all three sentence transformers and individual data tables}\label{tbl.baselinePerf_individual}
\begin{tabular*}{\tblwidth}{@{}LLRR@{}}
\toprule
Classification & Dataset & Bal. ACC & F$_1$-Score \\ 
\midrule 
\multirow{2}{*}{Binary} & Train & 1.000 & 1.000\\
& Test & 1.000 & 1.000\\
\midrule
\multirow{2}{*}{Three-class} & Train & 1.000 & 1.000\\
& Test & 0.865 & 0.844\\
\bottomrule
\end{tabular*}
\end{table}

\begin{table}[ht]
\caption{Peak classification performance of pooled baseline transformer classifications across all three sentence transformers and different sets of image features included}\label{tbl.baselinePerf_fused}
\begin{tabular*}{\tblwidth}{@{}LLRR@{}}
\toprule
Classification & Dataset & Bal. ACC & F$_1$-Score \\ 
\midrule 
\multirow{2}{*}{Binary} & Train & 1.000 & 1.000\\
& Test & 1.000 & 1.000\\
\midrule
\multirow{2}{*}{Three-class} & Train & 0.969 &  0.960\\
& Test & 0.887 & 0.868\\
\bottomrule
\end{tabular*}
\end{table}

\begin{table}[ht]
\caption{Peak performance of pooled baseline transformer classifications across different sets of image features included for individual sentence transformers}\label{tbl.baselineSentenceTransf}
\begin{tabular*}{\tblwidth}{@{}LLLRR@{}}
\toprule
Classification & Emb. Model & Dataset & Bal. ACC & F$_1$-Score \\ 
\midrule 
\multirow{6}{*}{Binary} & PubMedBert/ & Train & 1.000 & 1.000\\
& Small & Test & 1.000 & 1.000\\
\cmidrule{2-5} 
&\multirow{2}{*}{BioLORD} & Train & 1.000 & 1.000\\
&& Test & 0.985 & 0.987\\
\midrule
\multirow{6}{*}{Three-class} & \multirow{2}{*}{PubMedBert} & Train & 0.911 & 0.875\\
&& Test & 0.859 & 0.805\\
\cmidrule{2-5} 
&\multirow{2}{*}{BioLORD} & Train & 0.969 & 0.960\\
&& Test & 0.887 & 0.868\\
\cmidrule{2-5}
&\multirow{2}{*}{Small} & Train & 0.947 & 0.932\\
&& Test & 0.879 & 0.848\\
\bottomrule
\end{tabular*}
\end{table}

\begin{table}[ht]
\caption{Peak performance of pooled baseline transformer classifications across individual sentence transformers for different sets of image features included}\label{tbl.baselineImgMod}
\begin{tabular*}{\tblwidth}{@{}LLLRR@{}}
\toprule
Classification & Img. Modalities & Dataset & Bal. ACC & F$_1$-Score \\ 
\midrule 
\multirow{2}{*}{Binary} & \multirow{2}{*}{Any} & Train & 1.000 & 1.000\\
&& Test & 1.000 & 1.000\\
\midrule
\multirow{8}{*}{Three-class} & \multirow{2}{*}{None} & Train &  0.921 & 0.908\\
&& Test & 0.887 & 0.868\\
\cmidrule{2-5} 
&\multirow{2}{*}{ROI Volumes} & Train & 0.911 & 0.896\\
&& Test & 0.873 & 0.849\\
\cmidrule{2-5}
&\multirow{2}{*}{Prototype Sim.} & Train & 0.967 & 0.960\\
&& Test & 0.881 & 0.854\\
\cmidrule{2-5}
&\multirow{2}{*}{All} & Train & 0.969 & 0.960\\
&& Test & 0.876 & 0.848\\
\bottomrule
\end{tabular*}
\end{table}

\subsubsection{PatTree Performance}\label{patTreePerf}
Table \ref{tbl.best_PatTree} shows the peak performance achieved by the PatTree-based GNN models across all ablation studies. In the binary setting, nearly every choice for the ablation targets produced at least one model with perfect performance (bal. ACC and F$_1$ score of 1.0). The only configurations whose best model fell short of perfect performance were the BioLORD embeddings on the test set, the post\_emb\_sum data node vectors on the test set, and the Small embeddings on both training and test sets. However, even these reached bal. ACC and F$_1$ scores between 0.987 and 0.996. Eight models achieved perfect training performance, four of which also generalized to perfect test performance. The only ablation setting shared by all four was the use of PubMedBert embeddings. However, given the very high performance achieved with the other sentence transformers as well, we cannot conclude that PubMedBert is universally the best choice.

In the three-class setting, the best performance on both training and test sets was achieved by the model built on a PatTree using BioLORD embeddings, the post\_emb\_sum strategy for data node vectors, name embeddings (rather than zero vectors) for structure node property vectors, and prototype similarities as image features. This configuration reached bal. ACC scores of 99.3\% (training) and 98.5\% (test), with F$_1$ scores of 0.992 and 0.987, respectively.
By successfully applying the PatTree methodology to the selected ADNI-1 cohort, we demonstrate the feasibility of fully automated, early integration of heterogeneous, multimodal patient data for downstream data mining with GNNs. PatTree offers a way to build holistic patient representations that exploit the natural structure of patient journeys and clinical data system architectures, without requiring further modeling assumptions.

\begin{table}[ht]
\caption{Peak PatTree-based classification performance across all ablation studies}\label{tbl.best_PatTree}
\begin{tabular*}{\tblwidth}{@{}LLRR@{}}
\toprule
Classification & Dataset & Bal. ACC & F$_1$-Score \\ 
\midrule 
\multirow{2}{*}{Binary} & Train & 1.000 & 1.000\\
& Test & 1.000 & 1.000\\
\midrule
\multirow{2}{*}{Three-class} & Train & 0.993 & 0.992\\
& Test & 0.985 & 0.987\\
\bottomrule
\end{tabular*}
\end{table}

\subsubsection{Comparing PatTree to Baseline}\label{patTreeVSbaseline}
We compare the baseline transformers and our PatTree-based GNNs along two major dimensions: overall runtime and classification performance.

In terms of overall runtime, the baseline approach avoids most of the preprocessing steps, all graph-creation steps, and the computation of initial node vectors. The only additional step in the baseline approach is averaging table-specific classification results which is computationally negligible. As a result, its preparation time is far lower than that of the PatTree approach. This advantage carries over to training as well: the baseline transformers trained in a few seconds, compared to several hours for PatTree. Since both preparation and training are substantially faster for the baseline, its overall runtime is significantly shorter than that of PatTree.

In terms of classification performance, however, the picture is more favorable for PatTree. The two approaches performed equivalently in the binary setting, both achieving perfect classification, but PatTree clearly outperformed the baseline in the three-class setting, reaching a bal. ACC of 98.5\% compared to 88.7\% for the baseline. PatTree also offers two further practical advantages: it handles missing data and multiple patient visits natively, whereas the baseline requires an additional averaging strategy to accommodate them, and including image features improves PatTree's performance while degrading the baseline's.

\begin{table}[ht]
\caption{Direct comparison of peak performances of PatTree-based classifications and baseline transformer classifications on the held-out test set}\label{tbl.patTreeVSBaseline}
\begin{tabular*}{\tblwidth}{@{}LLRR@{}}
\toprule
Classification & Approach & Bal. ACC & F$_1$-Score \\ 
\midrule 
\multirow{2}{*}{Binary} & PatTree & 1.000 & 1.000\\
& Baseline Transformer & 1.000 & 1.000\\
\midrule
\multirow{2}{*}{Three-class} & PatTree & 0.985 & 0.987\\
& Baseline Transformer & 0.887 & 0.868\\
\bottomrule
\end{tabular*}
\end{table}

\subsection{Ablation Studies}
\subsubsection{Ablation Study 1: Sentence Transformer}
The first ablation study investigated the application of different sentence transformers to retrieve embeddings of node names and category terms. Results per sentence transformer are shown in table \ref{tbl.AblEmbMod}.

Across sentence transformers used to embed node names and category terms, the Small model was consistently less performant than the larger, domain-specific models. This observation is likely a result of its smaller embedding space, its lack of domain specificity, or both. PubMedBert and BioLORD performed similarly, with PubMedBert marginally ahead in the binary setting (test set) and BioLORD marginally ahead in the three-class setting. We therefore recommend using PubMedBert, BioLORD, or comparable domain-specific models when embedding node names and category terms for PatTree construction.

\begin{table}[ht]
\caption{Peak PatTree-based classification performance for different sentence transformers across all choices for other ablation targets}\label{tbl.AblEmbMod}
\begin{tabular*}{\tblwidth}{@{}LLLRR@{}}
\toprule
Classification & Emb. Model & Dataset & Bal. ACC & F$_1$-Score \\ 
\midrule 
\multirow{6}{*}{Binary} & \multirow{2}{*}{PubMedBert} & Train & 1.000 & 1.000\\
&& Test & 1.000 & 1.000\\
\cmidrule{2-5} 
&\multirow{2}{*}{BioLORD} & Train & 1.000 & 1.000\\
&& Test & 0.988 & 0.987\\
\cmidrule{2-5}
&\multirow{2}{*}{Small} & Train & 0.996 & 0.996\\
&& Test & 0.985 & 0.987\\
\midrule
\multirow{6}{*}{Three-class} & \multirow{2}{*}{PubMedBert} & Train & 0.983 & 0.982\\
&& Test & 0.972 & 0.966\\
\cmidrule{2-5} 
&\multirow{2}{*}{BioLORD} & Train & 0.993 & 0.992\\
&& Test & 0.985 & 0.987\\
\cmidrule{2-5}
&\multirow{2}{*}{Small} & Train & 0.924 & 0.915\\
&& Test & 0.922 & 0.905\\
\bottomrule
\end{tabular*}
\end{table}

\subsubsection{Ablation Study 2: Node Property Vectors of Data Nodes}
The second ablation study examined the influence of the strategy for combining node names with the data value when constructing the initial node property vector for data nodes in the PatTree. The results are shown in Table \ref{tbl.dataVecs}. The pre\_emb strategy combining the node name and data value prior to embedding performed poorly on the test set in the three-class setting. Besides the pre\_emb strategy, we tested two post\_emb combination strategies combining the node name and data value after embedding by summing or averaging the retrieved embeddings. These post\_emb strategies performed similarly, with post\_emb\_sum slightly weaker in the two-class setting and slightly stronger in the three-class setting. Given that the post\_emb strategies are also faster with respect to node name embedding and only slightly slower in node vector creation compared to the pre\_emb strategy, we recommend using post\_emb\_sum or post\_emb\_avg in PatTree construction. The choice between them can be treated as a hyperparameter to be tuned per project.

\begin{table}[ht]
\caption{Peak PatTree-based classification performance for different strategies for data node property vectors across all choices for other ablation targets}\label{tbl.dataVecs}
\begin{tabular*}{\tblwidth}{@{}LLLRR@{}}
\toprule
Classification & Comb. Strat. & Dataset & Bal. ACC & F$_1$-Score \\ 
\midrule 
\multirow{4}{*}{Binary} & Pre\_emb/ & Train & 1.000 & 1.000\\
& Post\_emb\_avg & Test & 1.000 & 1.000\\
\cmidrule{2-5} 
&\multirow{2}{*}{Post\_emb\_sum} & Train & 1.000 & 1.000\\
&& Test & 0.988 & 0.987\\
\midrule
\multirow{6}{*}{Three-class} & \multirow{2}{*}{Pre\_emb} & Train & 0.983 & 0.984\\
&& Test & 0.889 & 0.872\\
\cmidrule{2-5} 
&\multirow{2}{*}{Post\_emb\_sum} & Train & 0.993 & 0.992\\
&& Test & 0.985 & 0.987\\
\cmidrule{2-5}
&\multirow{2}{*}{Post\_emb\_avg} & Train & 0.981 & 0.982\\
&& Test & 0.956 & 0.956\\
\bottomrule
\end{tabular*}
\end{table}

\subsubsection{Ablation Study 3: Node Property Vectors of Structure Nodes}
The third ablation study compared using node name embeddings versus zero vectors as the node property vectors of structure nodes. The results are shown in Table \ref{tbl.strucVecs}. In the two-class setting, there was no difference between the two approaches, with all models achieving perfect performance. In the three-class setting, however, using name embeddings instead of zero vectors strictly—if only slightly—improved performance, indicating that the semantic information carried by these embeddings is meaningfully exploited by the model rather than adding noise. We recommend using name embeddings where the additional memory and computation cost is acceptable; in more resource-constrained settings, zero vectors remain a viable option with only a minor expected performance cost.

\begin{table}[ht]
\caption{Peak PatTree-based classification performance for different choices for structure node property vectors across all choices for other ablation targets}\label{tbl.strucVecs}
\begin{tabular*}{\tblwidth}{@{}LLLRR@{}}
\toprule
Classification & Structure Node Vec. & Dataset & Bal. ACC & F$_1$-Score \\ 
\midrule 
\multirow{2}{*}{Binary} & \multirow{2}{*}{Any} & Train & 1.000 & 1.000\\
&& Test & 1.000 & 1.000\\
\midrule
\multirow{4}{*}{Three-class} & \multirow{2}{*}{Zero} & Train & 0.981 & 0.982\\
&& Test & 0.967 & 0.974\\
\cmidrule{2-5} 
&\multirow{2}{*}{Embedding} & Train & 0.993 & 0.992\\
&& Test & 0.985 & 0.987\\
\bottomrule
\end{tabular*}
\end{table}

\subsubsection{Ablation Study 4: Image Features Included}
The fourth ablation study examined the influence of type and number of image features included. The results are shown in Table \ref{tbl.AblImgMod}. In the two-class setting, no differences emerged, with all configurations reaching perfect performance. In the three-class setting, the best performance was obtained when including Prototype Similarities, while adding ROI volumes reduced peak performance. One caveat is that Prototype Similarities are derived through supervised learning and could in principle leak class information; however, the baseline results on individual tables suggest this leakage, if present, is limited (c.f. section \ref{baselinePerf}). No firm conclusion can yet be drawn regarding the optimal choice of image features, and we suggest exploring more recent state-of-the-art feature extractors, such as 3DINO \cite{xu2025generalizable} or UNet architecture without skip connections (U-AE) such as DUNE \cite{barba2025dune}, in future work.

\begin{table}[ht]
\caption{Peak PatTree-based classification performance for different strategies for different imaging modalities included across all choices for other ablation targets}\label{tbl.AblImgMod}
\begin{tabular*}{\tblwidth}{@{}LLLRR@{}}
\toprule
Classification & Img. Modalities & Dataset & Bal. ACC & F$_1$-Score \\ 
\midrule 
\multirow{2}{*}{Binary} & \multirow{2}{*}{Any} & Train & 1.000 & 1.000\\
&& Test & 1.000 & 1.000\\
\midrule
\multirow{8}{*}{Three-class} & \multirow{2}{*}{None} & Train & 0.983 & 0.984\\
&& Test & 0.965 & 0.956\\
\cmidrule{2-5} 
&\multirow{2}{*}{ROI Volumes} & Train & 0.958 & 0.954\\
&& Test & 0.922 & 0.905\\
\cmidrule{2-5}
&\multirow{2}{*}{Prototype Sim.} & Train & 0.993 & 0.992\\
&& Test & 0.985 & 0.987\\
\cmidrule{2-5}
&\multirow{2}{*}{All} & Train & 0.983 & 0.982\\
&& Test & 0.972 & 0.966\\
\bottomrule
\end{tabular*}
\end{table}

\section{Conclusion}\label{concl}
The state-of-the-art classification performance results show that PatTree can be mined meaningfully demonstrating the suitability of PatTree as a holistic patient representation for medical classification tasks. Optimization of the PatTree structure and the GNN-based classification for different datasets remains as future work.

We proposed and implemented PatTree, a novel method for the automated structuring of medical real-world data that leverages the natural, intrinsic structure already present in such data (e.g., events, tables) rather than relying on tedious manual preparation and harmonization. The PatTree approach enables the construction of holistic, multimodal patient representations in a machine-understandable form at scale, without requiring assumptions about the data or upfront preparational effort. We demonstrated that direct classification on the PatTree representation is feasible using GNNs, achieving state-of-the-art ADNI classification performance with a bal. ACC of 98.5\% and an F$_1$ score of 0.987 in the three-class classification setting (AD vs. MCI vs. CN) on the held-out test set. In the two class classification setting (AD vs. CN), the PatTree approach reached perfect performance with a bal. ACC and an F$_1$ score of 1.0 which can mainly be attributed to the high correlation between the binary classification label (AD vs. CN) and the information contained by tables FNIHBC, MMSE, NEUROPATH, and CDR.

PatTree carries the potential to make medical real-world data usable from scratch, owing to its implicit handling of multiple modalities, missing values and modalities, and longitudinal data. These are three of the most reported challenges in working with medical real-world data in research. Additionally, PatTree supports holistic patient representations by integrating all available data for a given patient, while also enabling task-specific filtering of relevant information. This filtering can be achieved either through manual pruning of irrelevant tree branches by domain experts or through automated identification of relevant information via attention mechanisms or other branch-pruning strategies during task completion on the PatTree structure.

The analysis presented in this paper serves as a proof-of-concept, showing that GNN-based classification on PatTrees can reach state-of-the-art performance without resource-intensive manual data preparation. The anticipated robustness to missing values and modalities for individual patients, as well as the automated filtering capabilities, remain to be systematically investigated in future work.

In different ablations studies, we saw that different classification settings can benefit from different choices in PatTree construction. However, all examined options reached rather high classification performances, demonstrating that PatTree as a high-level idea works well and choices in the construction process act as targets for finetuning for optimal performance in individual projects.

Several directions for future work follow from our findings. First, automated filtering mechanisms should be implemented to identify and retain the most impactful subgraphs or branches within the PatTree structure. Second, the we could aim for improved PatTree classification performance through alternative message and update functions, the introduction of further semantic or residual edges, and systematic hyperparameter optimization for the GNN-based classifier. Third, mechanisms for explaining model decisions should be developed, for example by reading out subgraph embeddings at the event, table, or measurement node level and combining them via an attention mechanism that preserves traceability of which information contributed to a given decision. Finally, the generalizability of the approach should be evaluated in additional use cases with a special focus on examining PatTrees robustness to missing features and modalities. This also includes investigating the potential of including further modalities or applying different image feature extraction strategies.




\clearpage 




\section{}\label{}

\printcredits

\section{Funding}
We acknowledge funding by the Deutsche Forschungs-gemeinschaft (DFG, German Research Foundation) - project number: NFDI4DataScience (460234259). Data collection and sharing for the Alzheimer's Disease Neuroimaging Initiative (ADNI) is funded by the National Institute on Aging (National Institutes of Health Grant U19AG024904). The grantee organization is the Northern California Institute for Research and Education. In the past, ADNI has also received funding from the National Institute of Biomedical Imaging and Bioengineering, the Canadian Institutes of Health Research, and private sector contributions through the Foundation for the National Institutes of Health (FNIH) including generous contributions from the following: AbbVie, Alzheimer’s Association; Alzheimer’s Drug Discovery Foundation; Araclon Biotech; BioClinica, Inc.; Biogen; Bristol Myers Squibb Company; CereSpir, Inc.; Cogstate; Eisai Inc.; Elan Pharmaceuticals, Inc.; Eli Lilly and Company; EuroImmun; F. Hoffmann-La Roche Ltd and its affiliated company Genentech, Inc.; Fujirebio; GE Healthcare; IXICO Ltd.; Janssen Alzheimer Immunotherapy Research \& Development, LLC.; Johnson \& Johnson Pharmaceutical Research \& Development LLC.; Lumosity; Lundbeck; Merck \& Co., Inc.; Meso Scale Diagnostics, LLC.; NeuroRx Research; Neurotrack Technologies; Novartis Pharmaceuticals Corporation; Pfizer Inc.; Piramal Imaging; Servier; Takeda Pharmaceutical Company; and Transition Therapeutics.

\section{Declaration of Competing Interests}
The authors declare that they have no known competing financial interests or personal relationships that could have appeared to influence the work reported in this paper.

\section{Acknowledgements}
We thank our colleague Dr. Susanne Neufang for suggesting ADNI as the use case that served as the basis for evaluating PatTree and for supporting us in accesing and understanding the data. Furthermore, we thank Prof. Aleksandar Bojchevski for sharing his experience on GNNs which guided the design of our classification pipeline. Thirdly, we thank medical student Tim Dömges for his compilation of current KG approaches in the biomedical field which contributed to the related works section of this paper.

\section{Code and Data Availability Statement}

The source code implementing PatTree, including all data preprocessing, image-feature-extraction, string embedding, graph-construction, and classification models described in this manuscript, is publicly available at: \href{https://gitlab.git.nrw/uzk-cs/biomedical-informatics/gehrmann/pattree/-/tree/42f40f4f4af30e2f666519827ca32fc263aeb799/}{GitLab project PatTree}.

This link points to the exact commit used to produce the results reported in this paper; the project's main branch may continue to evolve after publication. The code is released under the Apache License 2.0.

The full experimental results underlying the analyses in this paper, provided as supplementary material, are available in \href{https://gitlab.git.nrw/uzk-cs/biomedical-informatics/gehrmann/pattree/-/tree/42f40f4f4af30e2f666519827ca32fc263aeb799/results/evaluation}{the results/evaluation directory of the same project}, including per-configuration baseline metrics, fused predictions, and PatTree evaluation outputs across embedding, modality, and classification-setting combinations.

The clinical and imaging data used in this study were obtained from the Alzheimer's Disease Neuroimaging Initiative (ADNI) and are not publicly redistributed as part of this repository due to ADNI's data-use agreement. Researchers wishing to access the underlying data should apply directly through ADNI (https://adni.loni.usc.edu/).

\bibliographystyle{unsrturl}

\bibliography{cas-refs}

@article{wang2024multilingual,
  title={Multilingual E5 Text Embeddings: A Technical Report},
  author={Wang, Liang and Yang, Nan and Huang, Xiaolong and Yang, Linjun and Majumder, Rangan and Wei, Furu},
  journal={arXiv preprint arXiv:2402.05672},
  year={2024},
  doi={10.48550/arXiv.2402.05672}
}

@article{remy-etal-2023-biolord,
    author = {Remy, François and Demuynck, Kris and Demeester, Thomas},
    title = "{BioLORD-2023: semantic textual representations fusing large language models and clinical knowledge graph insights}",
    journal = {J. Am. Med. Inform. Assoc.},
    pages = {ocae029},
    year = {2024},
    month = {02},
    issn = {1527-974X},
    doi = {10.1093/jamia/ocae029}
}

@article{gu2021domain,
  title={Domain-specific language model pretraining for biomedical natural language processing},
  author={Gu, Yu and Tinn, Robert and Cheng, Hao and Lucas, Michael and Usuyama, Naoto and Liu, Xiaodong and Naumann, Tristan and Gao, Jianfeng and Poon, Hoifung},
  journal={ACM Trans. Comput. Healthc.},
  volume={3},
  number={1},
  pages={1--23},
  year={2021},
  publisher={ACM New York, NY},
  doi={10.1145/3458754}
}

@article{abuhantash2024comorbidity,
  title={Comorbidity-based framework for Alzheimer’s disease classification using graph neural networks},
  author={Abuhantash, Ferial and Abu Hantash, Mohd Khalil and AlShehhi, Aamna},
  journal={Sci. Rep.},
  volume={14},
  number={1},
  pages={21061},
  year={2024},
  publisher={Nature Publishing Group UK London},
  doi={10.1038/s41598-024-72321-2}
}

@article{sokolova2009systematic,
  title={A systematic analysis of performance measures for classification tasks},
  author={Sokolova, Marina and Lapalme, Guy},
  journal={Inf. Process. Manag.},
  volume={45},
  number={4},
  pages={427--437},
  year={2009},
  publisher={Elsevier},
  doi={10.1016/j.ipm.2009.03.002}
}

@inbook{valiente2002treesandgraphs,
    author = {Valiente, Gabriel},
    title = {Algorithms on trees and graphs},
    publisher = {Springer},
    year = {2002},
    volume={112},
    chapter = {1.1},
    pages = {3-19},
    doi={10.1007/978-3-030-81885-2}
}

@article{barba2025dune,
  title={DUNE: a versatile neuroimaging encoder captures brain complexity across 3 major diseases: cancer, dementia, and schizophrenia},
  author={Barba, Thomas and Bagley, Bryce A and Steyaert, Sandra and Carrillo-Perez, Francisco and Sad{\'e}e, Christoph and Iv, Michael and Gevaert, Olivier},
  journal={GigaScience},
  volume={14},
  pages={giaf116},
  year={2025},
  publisher={Oxford University Press},
  doi={10.1093/gigascience/giaf116}
}

@article{xu2025generalizable,
  title={A generalizable 3D framework and model for self-supervised learning in medical imaging},
  author={Xu, Tony and Hosseini, Sepehr and Anderson, Chris and Rinaldi, Anthony and Krishnan, Rahul G and Martel, Anne L and Goubran, Maged},
  journal={NPJ Digit. Med.},
  volume={8},
  number={1},
  pages={639},
  year={2025},
  publisher={Nature Publishing Group UK London},
  doi={10.1038/s41746-025-02035-w}
}

@Article{diagnostics16121755,
AUTHOR = {Xu, Jiayuan and Costen, Fumie},
TITLE = {Machine Learning-Based Multiclass Classification of Cognitive Stages Using Plasma Biomarkers, Clinical Assessments, and Genetic Features: A Repeated, Nested Cross-Validation Study in ADNI with External Evaluation in CNTN},
JOURNAL = {Diagnostics},
VOLUME = {16},
YEAR = {2026},
NUMBER = {12},
ARTICLE-NUMBER = {1755},
PubMedID = {42351415},
ISSN = {2075-4418},
DOI = {10.3390/diagnostics16121755}
}

@article{zheng2025scoping,
  title={A scoping review of self-supervised representation learning for clinical decision making using EHR categorical data},
  author={Zheng, Yuanyuan and Bensahla, Adel and Bjelogrlic, Mina and Zaghir, Jamil and Turbe, Hugues and Bednarczyk, Lydie and Gaudet-Blavignac, Christophe and Ehrsam, Julien and Marchand-Maillet, St{\'e}phane and Lovis, Christian},
  journal={NPJ Digit. Med.},
  volume={8},
  number={1},
  pages={362},
  year={2025},
  doi={10.1038/s41746-025-01692-1}
}

@inproceedings{choi2020learning,
  title={Learning the graphical structure of electronic health records with graph convolutional transformer},
  author={Choi, Edward and Xu, Zhen and Li, Yujia and Dusenberry, Michael and Flores, Gerardo and Xue, Emily and Dai, Andrew},
  booktitle={Proceedings of the AAAI conference on artificial intelligence},
  volume={34},
  number={01},
  pages={606--613},
  year={2020},
  doi={10.1609/aaai.v34i01.5400}
}

@inproceedings{choi2017gram,
  title={GRAM: graph-based attention model for healthcare representation learning},
  author={Choi, Edward and Bahadori, Mohammad Taha and Song, Le and Stewart, Walter F and Sun, Jimeng},
  booktitle={Proceedings of the 23rd ACM SIGKDD international conference on knowledge discovery and data mining},
  pages={787--795},
  year={2017},
  doi={10.1145/3097983.3098126}
}

@article{sun2023scoping,
  title={A scoping review on multimodal deep learning in biomedical images and texts},
  author={Sun, Zhaoyi and Lin, Mingquan and Zhu, Qingqing and Xie, Qianqian and Wang, Fei and Lu, Zhiyong and Peng, Yifan},
  journal={J. Biomed. Inform.},
  volume={146},
  pages={104482},
  year={2023},
  publisher={Elsevier},
  doi={10.1016/j.jbi.2023.104482}
}

@article{finster2025common,
  title={Common data models and data standards for tabular health data: a systematic review},
  author={Finster, Melissa and Wenzel, Markus and Taghizadeh, Elham},
  journal={BMC Med. Inform. Decis. Mak.},
  volume={25},
  number={1},
  pages={422},
  year={2025},
  publisher={Springer},
  doi={10.1186/s12911-025-03267-2}
}

@article{ayaz2021fast,
  title={The Fast Health Interoperability Resources (FHIR) standard: systematic literature review of implementations, applications, challenges and opportunities},
  author={Ayaz, Muhammad and Pasha, Muhammad F and Alzahrani, Mohammed Y and Budiarto, Rahmat and Stiawan, Deris},
  journal={JMIR Med. Inform.},
  volume={9},
  number={7},
  pages={e21929},
  year={2021},
  publisher={JMIR Publications Toronto, Canada},
  doi={10.2196/21929}
}

@article{klann2018web,
  title={Web services for data warehouses: OMOP and PCORnet on i2b2},
  author={Klann, Jeffrey G and Phillips, Lori C and Herrick, Christopher and Joss, Matthew AH and Wagholikar, Kavishwar B and Murphy, Shawn N},
  journal={J. Am. Med. Inform. Assoc.},
  volume={25},
  number={10},
  pages={1331--1338},
  year={2018},
  publisher={Oxford University Press},
  doi={10.1093/jamia/ocy093}
}

@article{hripcsak2015observational,
  title={Observational Health Data Sciences and Informatics (OHDSI): opportunities for observational researchers},
  author={Hripcsak, George and Duke, Jon D and Shah, Nigam H and Reich, Christian G and Huser, Vojtech and Schuemie, Martijn J and Suchard, Marc A and Park, Rae Woong and Wong, Ian Chi Kei and Rijnbeek, Peter R and others},
  journal={Stud. Health Technol. Inform.},
  volume={216},
  pages={574},
  year={2015},
  doi={10.3233/978-1-61499-564-7-574}
}

@article{overhage2012validation,
  title={Validation of a common data model for active safety surveillance research},
  author={Overhage, J Marc and Ryan, Patrick B and Reich, Christian G and Hartzema, Abraham G and Stang, Paul E},
  journal={J. Am. Med. Inform. Assoc.},
  volume={19},
  number={1},
  pages={54--60},
  year={2012},
  publisher={BMJ Group BMA House, Tavistock Square, London, WC1H 9JR},
  doi={10.1136/amiajnl-2011-000376}
}

@article{pellegrini2018machine,
  title={Machine learning of neuroimaging for assisted diagnosis of cognitive impairment and dementia: a systematic review},
  author={Pellegrini, Enrico and Ballerini, Lucia and Hernandez, Maria del C Valdes and Chappell, Francesca M and Gonz{\'a}lez-Castro, Victor and Anblagan, Devasuda and Danso, Samuel and Mu{\~n}oz-Maniega, Susana and Job, Dominic and Pernet, Cyril and others},
  journal={Alzheimer's Dement.: Diagn. Assess. Dis. Monit.},
  volume={10},
  number={1},
  pages={519--535},
  year={2018},
  publisher={Wiley Online Library},
  doi={10.1016/j.dadm.2018.07.004}
}

@article{tanveer2020machine,
  title={Machine learning techniques for the diagnosis of Alzheimer’s disease: A review},
  author={Tanveer, Muhammad and Richhariya, Bharat and Khan, Riyaj Uddin and Rashid, Ashraf Haroon and Khanna, Pritee and Prasad, Mukesh and Lin, Chin-Teng},
  journal={ACM Trans. Multimed. Comput. Commun. Appl.},
  volume={16},
  number={1s},
  pages={1--35},
  year={2020},
  publisher={ACM New York, NY, USA},
  doi={10.1145/3344998}
}

@article{diogo2022early,
  title={Early diagnosis of Alzheimer’s disease using machine learning: a multi-diagnostic, generalizable approach},
  author={Diogo, Vasco S{\'a} and Ferreira, Hugo Alexandre and Prata, Diana and Alzheimer’s Disease Neuroimaging Initiative},
  journal={Alzheimer's Res. Ther.},
  volume={14},
  number={1},
  pages={107},
  year={2022},
  publisher={Springer},
  doi={10.1186/s13195-022-01047-y}
}

@article{mieling2026predicting,
  title={Predicting the progression of MCI and Alzheimer’s disease on structural brain integrity and other features with machine learning},
  author={Mieling, Marthe and Yousuf, Mushfa and Bunzeck, Nico},
  journal={GeroScience},
  volume={48},
  number={1},
  pages={463--487},
  year={2026},
  publisher={Springer},
  doi={10.1007/s11357-025-01626-5}
}

@article{zhou2025deep,
  title={A deep learning model for early diagnosis of alzheimer’s disease combined with 3D CNN and video Swin transformer},
  author={Zhou, Juan and Wei, Yiming and Li, Xiong and Zhou, Weiqiang and Tao, Ruiyang and Hua, Yi and Liu, Hongwei},
  journal={Sci. Rep.},
  volume={15},
  number={1},
  pages={23311},
  year={2025},
  publisher={Nature Publishing Group UK London},
  doi={10.1038/s41598-025-05568-y}
}

@article{hussain2025alzformer,
  title={AlzFormer: Multi-modal framework for alzheimer’s classification using MRI and graph-embedded demographics guided by adaptive attention gating},
  author={Hussain, Sayyed Shahid and Degang, Xu and Shah, Pir Masoom and Khan, Hikmat and Zeb, Adnan},
  journal={Comput. Med. Imaging Graph.},
  pages={102638},
  year={2025},
  publisher={Elsevier},
  doi={10.1016/j.compmedimag.2025.102638}
}

@article{wen2020convolutional,
  title={Convolutional neural networks for classification of Alzheimer's disease: Overview and reproducible evaluation},
  author={Wen, Junhao and Thibeau-Sutre, Elina and Diaz-Melo, Mauricio and Samper-Gonz{\'a}lez, Jorge and Routier, Alexandre and Bottani, Simona and Dormont, Didier and Durrleman, Stanley and Burgos, Ninon and Colliot, Olivier and others},
  journal={Med. Image Anal.},
  volume={63},
  pages={101694},
  year={2020},
  publisher={Elsevier},
  doi={10.1016/j.media.2020.101694}
}

@article{gorishniy2021revisiting,
  title={Revisiting deep learning models for tabular data},
  author={Gorishniy, Yury and Rubachev, Ivan and Khrulkov, Valentin and Babenko, Artem},
  journal={Adv. Neural Inf. Pocess. Syst.},
  volume={34},
  pages={18932--18943},
  year={2021},
  doi={10.48550/arXiv.2106.11959}
}

@article{brody2021attentive,
  title={How attentive are graph attention networks?},
  author={Brody, Shaked and Alon, Uri and Yahav, Eran},
  journal={arXiv preprint arXiv:2105.14491},
  year={2021},
  doi={10.48550/arXiv.2105.14491}
}

@inproceedings{velivckovic2018graph,
  title={Graph attention networks},
  author={Veli{\v{c}}kovi{\'c}, Petar and Cucurull, Guillem and Casanova, Arantxa and Romero, Adriana and Lio, Pietro and Bengio, Yoshua and others},
  booktitle={International conference on learning representations},
  volume={6},
  number={2},
  year={2018},
  organization={Ithaca},
  doi={10.17863/CAM.48429}
}

@inproceedings{wang_2018-07_NovelMultimodalMRI,
	location = {Honolulu, {HI}},
	title = {A Novel Multimodal {MRI} Analysis for Alzheimer's Disease Based on Convolutional Neural Network},
	isbn = {978-1-5386-3646-6},
	doi = {10.1109/EMBC.2018.8512372},
	eventtitle = {2018 40th Annual International Conference of the {IEEE} Engineering in Medicine and Biology Society ({EMBC})},
	pages = {754--757},
	booktitle = {2018 40th Annual International Conference of the {IEEE} Engineering in Medicine and Biology Society ({EMBC})},
	publisher = {{IEEE}},
	author = {Wang, Yan and Yang, Yanwu and Guo, Xin and Ye, Chenfei and Gao, Na and Fang, Yuan and Ma, Heather T.},
	date = {2018-07},
}

@article{le2020challenges,
  title={The challenges in data integration--heterogeneity and complexity in clinical trials and patient registries of Systemic Lupus Erythematosus},
  author={Le Sueur, Helen and Bruce, Ian N and Geifman, Nophar and Masterplans Consortium},
  journal={BMC Med. Res. Methodol.},
  volume={20},
  number={1},
  pages={164},
  year={2020},
  publisher={Springer},
  doi={10.1186/s12874-020-01057-0}
}

@article{goyal2025named,
  title={Named entity recognition and relationship extraction for biomedical text: A comprehensive survey, recent advancements, and future research directions},
  author={Goyal, Nandita and Singh, Navdeep},
  journal={Neurocomputing},
  volume={618},
  pages={129171},
  year={2025},
  publisher={Elsevier},
  doi={10.1016/j.neucom.2024.129171}
}

@article{sachdeva2022using,
  title={Using knowledge graph structures for semantic interoperability in electronic health records data exchanges},
  author={Sachdeva, Shelly and Bhalla, Subhash},
  journal={Inf.},
  volume={13},
  number={2},
  pages={52},
  year={2022},
  publisher={MDPI},
  doi={10.3390/info13020052}
}

@article{frau2025connecting,
  title={Connecting electronic health records to a biomedical knowledge graph to link clinical phenotypes and molecular endotypes in atopic dermatitis},
  author={Frau, Francesca and Loustalot, Paul and T{\"o}rnqvist, Margaux and Temam, Nina and Cupe, Jean and Montmerle, Martin and Aug{\'e}, Franck},
  journal={Sci. Rep.},
  volume={15},
  number={1},
  pages={3082},
  year={2025},
  publisher={Nature Publishing Group UK London},
  doi={10.1038/s41598-024-78794-5}
}

@article{gehrmann2023prevents,
  title={What prevents us from reusing medical real-world data in research},
  author={Gehrmann, Julia and Herczog, Edit and Decker, Stefan and Beyan, Oya},
  journal={Sci. Data},
  volume={10},
  number={1},
  pages={459},
  year={2023},
  publisher={Nature Publishing Group UK London},
  doi = "10.1038/s41597-023-02361-2"
}

@incollection{morris2022graph,
  title={Graph neural networks: Graph classification},
  author={Morris, Christopher},
  booktitle={Graph Neural Networks: Foundations, Frontiers, and Applications},
  pages={179--193},
  year={2022},
  publisher={Springer},
  doi={10.1007/978-981-16-6054-2_9}
}

@inproceedings{grohe2020word2vec,
  title={word2vec, node2vec, graph2vec, x2vec: Towards a theory of vector embeddings of structured data},
  author={Grohe, Martin},
  booktitle={proceedings of the 39th ACM SIGMOD-SIGACT-SIGAI symposium on principles of database systems},
  pages={1--16},
  year={2020},
  doi={10.1145/3375395.3387641}
}

@article{gehrmann2024early,
  title={Early Multimodal Data Integration for Data-Driven Medical Research--A Scoping Review},
  author={Gehrmann, Julia and Beyan, Oya},
  journal={Ger. Med. Data Sci. 2024},
  pages={49--58},
  year={2024},
  publisher={IOS Press},
  doi={10.3233/SHTI240837}
}

@article{goryawala2015inclusion,
  title={Inclusion of neuropsychological scores in atrophy models improves diagnostic classification of Alzheimer’s disease and mild cognitive impairment},
  author={Goryawala, Mohammed and Zhou, Qi and Barker, Warren and Loewenstein, David A and Duara, Ranjan and Adjouadi, Malek},
  journal={Comput. Intell. Neurosci.},
  volume={2015},
  number={1},
  pages={865265},
  year={2015},
  publisher={Wiley Online Library},
  doi={10.1155/2015/865265}
}

@inproceedings{salem2025transformer,
  title={Transformer Models in Natural Language Processing: A Comprehensive Review and Prospects for Future Development},
  author={Salem, Maha and Mohamed, Azza and Shaalan, Khaled},
  booktitle={International Conference on Advanced Intelligent Systems and Informatics},
  pages={463--472},
  year={2025},
  organization={Springer},
  doi={10.1007/978-3-031-81308-5_42}
}

@inproceedings{zhang2024comparative,
  title={A comparative study of One-Hot, TF-IDF, and Word2Vec for Classifying Illegal Advertising Texts},
  author={Zhang, Yuedan and He, Lingmin and Zhang, Yunpeng and Zhao, Panzhi and Zhang, Bole and Cheng, Fang},
  booktitle={Proceedings of the 2024 8th International Conference on Natural Language Processing and Information Retrieval},
  pages={82--86},
  year={2024},
  doi={10.1145/3711542.3711586}
}

@article{chaki2026deep,
  title={The Deep Learning Revolution in Neuroimaging: Insights from a Bibliometric Analysis (2014--2024)},
  author={Chaki, Jyotismita and Deshpande, Gopikrishna},
  journal={Neuroinformatics},
  volume={24},
  number={2},
  pages={16},
  year={2026},
  publisher={Springer},
  doi={10.1007/s12021-026-09775-4}
}

@article{mayerhoefer2020introduction,
  title={Introduction to radiomics},
  author={Mayerhoefer, Marius E and Materka, Andrzej and Langs, Georg and H{\"a}ggstr{\"o}m, Ida and Szczypi{\'n}ski, Piotr and Gibbs, Peter and Cook, Gary},
  journal={J. Nucl. Med.},
  volume={61},
  number={4},
  pages={488--495},
  year={2020},
  publisher={Society of Nuclear Medicine},
  doi={10.2967/jnumed.118.222893}
}

@article{dehbozorgi2025comparative,
  title={A comparative study of statistical, radiomics, and deep learning feature extraction techniques for medical image classification in optical and radiological modalities},
  author={Dehbozorgi, Pegah and Ryabchykov, Oleg and Bocklitz, Thomas W},
  journal={Comput. Biol. Med.},
  volume={187},
  pages={109768},
  year={2025},
  publisher={Elsevier},
  doi={10.1016/j.compbiomed.2025.109768}
}

@article{hallur2025image,
  title={Image feature extraction techniques: A comprehensive review},
  author={Hallur, Sudhakar and Gavade, Anil},
  journal={Franklin Open},
  pages={100366},
  year={2025},
  publisher={Elsevier},
  doi={10.1016/j.fraope.2025.100366}
}

@article{mueller2005adni,
  title={Ways toward an early diagnosis in Alzheimer’s disease: the Alzheimer’s Disease Neuroimaging Initiative (ADNI)},
  author={Mueller, Susanne G and Weiner, Michael W and Thal, Leon J and Petersen, Ronald C and Jack, Clifford R and Jagust, William and Trojanowski, John Q and Toga, Arthur W and Beckett, Laurel},
  journal={Alzheimer's Dement.},
  volume={1},
  number={1},
  pages={55--66},
  year={2005},
  publisher={Elsevier},
  doi={10.1016/j.jalz.2005.06.003}
}

@inproceedings{nauta2023pip,
  title={Pip-net: Patch-based intuitive prototypes for interpretable image classification},
  author={Nauta, Meike and Schl{\"o}tterer, J{\"o}rg and Van Keulen, Maurice and Seifert, Christin},
  booktitle={Proceedings of the IEEE/CVF conference on computer vision and pattern recognition},
  pages={2744--2753},
  year={2023},
  doi={10.1109/CVPR52729.2023.00269}
}

@article{ebrahimi2021convolutional,
  title={Convolutional neural networks for Alzheimer’s disease detection on MRI images},
  author={Ebrahimi, Amir and Luo, Suhuai and Disease Neuroimaging Initiative, for the Alzheimer’s},
  journal={J. Med. Imaging},
  volume={8},
  number={2},
  pages={024503--024503},
  year={2021},
  publisher={Society of Photo-Optical Instrumentation Engineers},
  doi={10.1117/1.JMI.8.2.024503}
}

@article{vaswani2017attention,
  title={Attention is all you need},
  author={Vaswani, Ashish and Shazeer, Noam and Parmar, Niki and Uszkoreit, Jakob and Jones, Llion and Gomez, Aidan N and Kaiser, {\L}ukasz and Polosukhin, Illia},
  journal={Adv. Neural Inf. Process. Syst.},
  volume={30},
  year={2017},
  doi={10.48550/arXiv.1706.03762}
}

@article{mikolov2013efficient,
  title={Efficient estimation of word representations in vector space},
  author={Mikolov, Tomas and Chen, Kai and Corrado, Greg and Dean, Jeffrey},
  journal={arXiv preprint arXiv:1301.3781},
  year={2013},
  doi={10.48550/arXiv.1301.3781}
}

@book{hogan2022knowledge,
	address = {Cham},
	series = {Synthesis {Lectures} on {Data}, {Semantics}, and {Knowledge}},
	title = {Knowledge Graphs},
	copyright = {https://www.springernature.com/gp/researchers/text-and-data-mining},
	isbn = {978-3-031-00790-3 978-3-031-01918-0},
	language = {en},
	publisher = {Springer International Publishing},
	author = {Hogan, Aidan and Gutierrez, Claudio and Cochez, Michael and Melo, Gerard De and Kirrane, Sabrina and Polleres, Axel and Navigli, Roberto and Ngomo, Axel-Cyrille Ngonga and Rashid, Sabbir M. and Schmelzeisen, Lukas and Staab, Steffen and Blomqvist, Eva and d’Amato, Claudia and Gayo, José Emilio Labra and Neumaier, Sebastian and Rula, Anisa and Sequeda, Juan and Zimmermann, Antoine},
	year = {2022},
	doi = {10.1007/978-3-031-01918-0}
}

@inproceedings{de2024pipnet3d,
  title={Pipnet3d: Interpretable detection of alzheimer in mri scans},
  author={De Santi, Lisa Anita and Schl{\"o}tterer, J{\"o}rg and Scheschenja, Michael and Wessendorf, Joel and Nauta, Meike and Positano, Vincenzo and Seifert, Christin},
  booktitle={International Conference on Medical Image Computing and Computer-Assisted Intervention},
  pages={69--78},
  year={2024},
  organization={Springer},
  doi={10.1007/978-3-031-77610-6_7}
}

@inproceedings{bloch2021comparison,
  title={Comparison of Automated Volume Extraction With FreeSurfer and FastSurfer for Early Alzheimer's Disease Detection With Machine Learning},
  author={Bloch, Louise and Friedrich, Christoph M},
  booktitle={2021 IEEE 34th International Symposium on Computer-Based Medical Systems (CBMS)},
  pages={113--118},
  year={2021},
  organization={IEEE},
  doi={10.1109/CBMS52027.2021.00096}
}

@article{henschel2020fastsurfer,
  title={Fastsurfer-a fast and accurate deep learning based neuroimaging pipeline},
  author={Henschel, Leonie and Conjeti, Sailesh and Estrada, Santiago and Diers, Kersten and Fischl, Bruce and Reuter, Martin},
  journal={NeuroImage},
  volume={219},
  pages={117012},
  year={2020},
  publisher={Elsevier},
  doi={10.1016/j.neuroimage.2020.117012}
}

@article{yang2025large,
  title={Large language model--driven knowledge graph construction in sepsis care using multicenter clinical databases: Development and Usability Study},
  author={Yang, Hao and Li, Jiaxi and Zhang, Chi and Sierra, Alejandro Pazos and Shen, Bairong},
  journal={J. Med. Internet Res.},
  volume={27},
  pages={e65537},
  year={2025},
  publisher={JMIR Publications Toronto, Canada},
  doi={10.2196/65537}
}

@article{sengupta2025medaka,
  title={MEDAKA: Construction of Biomedical Knowledge Graphs Using Large Language Models},
  author={Sengupta, Asmita and Selby, David Antony and Vollmer, Sebastian Josef and Gro{\ss}mann, Gerrit},
  journal={arXiv preprint arXiv:2509.26128},
  year={2025},
  doi={10.48550/arXiv.2509.26128}
}

@article{seneviratne2021personal,
  title={Personal health knowledge graph for clinically relevant diet recommendations},
  author={Seneviratne, Oshani and Harris, Jonathan and Chen, Ching-Hua and McGuinness, Deborah L},
  journal={arXiv preprint arXiv:2110.10131},
  year={2021},
  doi={10.48550/arXiv.2110.10131}
}

@article{bloor2023towards,
  title={Towards a Personal Health Knowledge Graph Framework for Patient Monitoring},
  author={Bloor, Daniel and Ugwuoke, Nnamdi and Taylor, David and Lewis, Keir and Mur, Luis and Lu, Chuan},
  journal={arXiv preprint arXiv:2311.06524},
  year={2023},
  doi={10.48550/arXiv.2311.06524}
}

@inproceedings{guluzade2021demographic,
  title={Demographic aware probabilistic medical knowledge graph embeddings of electronic medical records},
  author={Guluzade, Aynur and Kacupaj, Endri and Maleshkova, Maria},
  booktitle={International Conference on Artificial Intelligence in Medicine},
  pages={408--417},
  year={2021},
  organization={Springer},
  doi={10.1007/978-3-030-77211-6_48}
}

@article{li2020real,
  title={Real-world data medical knowledge graph: construction and applications},
  author={Li, Linfeng and Wang, Peng and Yan, Jun and Wang, Yao and Li, Simin and Jiang, Jinpeng and Sun, Zhe and Tang, Buzhou and Chang, Tsung-Hui and Wang, Shenghui and others},
  journal={Artif. Intell. Med.},
  volume={103},
  pages={101817},
  year={2020},
  publisher={Elsevier},
  doi={10.1016/j.artmed.2020.101817}
}

@inproceedings{aldughayfiq2023capturing,
  title={Capturing semantic relationships in electronic health records using knowledge graphs: An implementation using mimic iii dataset and graphdb},
  author={Aldughayfiq, Bader and Ashfaq, Farzeen and Jhanjhi, NZ and Humayun, Mamoona},
  booktitle={Healthcare},
  volume={11},
  number={12},
  pages={1762},
  year={2023},
  organization={MDPI},
  doi={10.3390/healthcare11121762}
}

@article{chandak2023building,
  title={Building a knowledge graph to enable precision medicine},
  author={Chandak, Payal and Huang, Kexin and Zitnik, Marinka},
  journal={Sci. Data},
  volume={10},
  number={1},
  pages={67},
  year={2023},
  publisher={Nature Publishing Group UK London},
  doi={10.1038/s41597-023-01960-3}
}

@article{morris2023scalable,
  title={The scalable precision medicine open knowledge engine (SPOKE): a massive knowledge graph of biomedical information},
  author={Morris, John H and Soman, Karthik and Akbas, Rabia E and Zhou, Xiaoyuan and Smith, Brett and Meng, Elaine C and Huang, Conrad C and Cerono, Gabriel and Schenk, Gundolf and Rizk-Jackson, Angela and others},
  journal={Bioinform.},
  volume={39},
  number={2},
  pages={btad080},
  year={2023},
  publisher={Oxford University Press},
  doi={10.1093/bioinformatics/btad080}
}

@article{boehm2022harnessing,
  title={Harnessing multimodal data integration to advance precision oncology},
  author={Boehm, Kevin M and Khosravi, Pegah and Vanguri, Rami and Gao, Jianjiong and Shah, Sohrab P},
  journal={Nat. Rev. Cancer},
  volume={22},
  number={2},
  pages={114--126},
  year={2022},
  publisher={Nature Publishing Group UK London},
  doi={10.1038/s41568-021-00408-3}
}

@article{bokade2021cross,
  title={A cross-disciplinary comparison of multimodal data fusion approaches and applications: Accelerating learning through trans-disciplinary information sharing},
  author={Bokade, Rohit and Navato, Alfred and Ouyang, Ruilin and Jin, Xiaoning and Chou, Chun-An and Ostadabbas, Sarah and Mueller, Amy V},
  journal={Expert Syst. Appl.},
  volume={165},
  pages={113885},
  year={2021},
  publisher={Elsevier},
  doi={10.1016/j.eswa.2020.113885}
}

@article{kline2022multimodal,
  title={Multimodal machine learning in precision health: A scoping review},
  author={Kline, Adrienne and Wang, Hanyin and Li, Yikuan and Dennis, Saya and Hutch, Meghan and Xu, Zhenxing and Wang, Fei and Cheng, Feixiong and Luo, Yuan},
  journal={NPJ Digit. Med.},
  volume={5},
  number={1},
  pages={171},
  year={2022},
  publisher={Nature Publishing Group UK London},
  doi={10.1038/s41746-022-00712-8}
}

@article{behrad2022overview,
  title={An overview of deep learning methods for multimodal medical data mining},
  author={Behrad, Fatemeh and Abadeh, Mohammad Saniee},
  journal={Expert Syst. Appl.},
  volume={200},
  pages={117006},
  year={2022},
  publisher={Elsevier},
  doi={10.1016/j.eswa.2022.117006}
}

@article{lee2017medical,
  title={Medical big data: promise and challenges},
  author={Lee, Choong Ho and Yoon, Hyung-Jin},
  journal={Kidney Res. Clin. Pract.},
  volume={36},
  number={1},
  pages={3},
  year={2017},
  doi={10.23876/j.krcp.2017.36.1.3}
}

@incollection{mazein2024medax,
  title={MeDaX: A Knowledge Graph on FHIR},
  author={Mazein, Ilya and Gebhardt, Tom and Zinkewitz, Felix and Michaelis, Lea and Braun, Sarah and Waltemath, Dagmar and Henkel, Ron and Wodke, Judith AH},
  booktitle={Digital Health and Informatics Innovations for Sustainable Health Care Systems: Proceedings of MIE 2024},
  pages={367--371},
  year={2024},
  publisher={SAGE Publications 1 Oliver's Yard, 55 City Road, London, EC1Y 1SP},
  doi={10.3233/SHTI240423}
}

@article{xiao2022fhir,
  title={FHIR-Ontop-OMOP: Building clinical knowledge graphs in FHIR RDF with the OMOP Common data Model},
  author={Xiao, Guohui and Pfaff, Emily and Prud'hommeaux, Eric and Booth, David and Sharma, Deepak K and Huo, Nan and Yu, Yue and Zong, Nansu and Ruddy, Kathryn J and Chute, Christopher G and others},
  journal={J. Biomed. Inform.},
  volume={134},
  pages={104201},
  year={2022},
  publisher={Elsevier},
  doi={10.1016/j.jbi.2022.104201}
}

@article{kang2024evolution,
  title={Evolution of a graph model for the OMOP common data model},
  author={Kang, Mengjia and Alvarado-Guzman, Jose A and Rasmussen, Luke V and Starren, Justin B},
  journal={Appl. Clin. Inform.},
  volume={15},
  number={05},
  pages={1056--1065},
  year={2024},
  publisher={Georg Thieme Verlag KG},
  doi={10.1055/s-0044-1791487}
}

@incollection{chytas2024mapping,
  title={Mapping OMOP-CDM to RDF: bringing real-world-data to the semantic web realm},
  author={Chytas, Achilleas and Bassileiades, Nick and Natsiavas, Pantelis},
  booktitle={Digital Health and Informatics Innovations for Sustainable Health Care Systems},
  pages={1406--1410},
  year={2024},
  publisher={IOS Press},
  doi={10.3233/SHTI240674}
}

@inproceedings{gori2005new,
  title={A new model for learning in graph domains},
  author={Gori, Marco and Monfardini, Gabriele and Scarselli, Franco},
  booktitle={Proceedings. 2005 IEEE international joint conference on neural networks, 2005.},
  volume={2},
  pages={729--734},
  year={2005},
  organization={IEEE},
  doi={10.1109/IJCNN.2005.1555942}
}

@article{qiu2022multimodal,
  title={Multimodal deep learning for Alzheimer’s disease dementia assessment},
  author={Qiu, Shangran and Miller, Matthew I and Joshi, Prajakta S and Lee, Joyce C and Xue, Chonghua and Ni, Yunruo and Wang, Yuwei and De Anda-Duran, Ileana and Hwang, Phillip H and Cramer, Justin A and others},
  journal={Nat. Commun.},
  volume={13},
  number={1},
  pages={3404},
  year={2022},
  publisher={Nature Publishing Group UK London},
  doi={10.1038/s41467-022-31037-5}
}

@article{shaikh2025deep,
  title={Deep learning for Alzheimer’s disease: advances in classification, segmentation, subtyping, and explainability},
  author={Shaikh, Mohammed Rizwan and Jeyabose, Andrew and Arjunan, R Vijaya},
  journal={Biomed. Eng. Online},
  volume={24},
  number={1},
  pages={150},
  year={2025},
  publisher={Springer},
  doi={10.1186/s12938-025-01482-6}
}

@article{ali2025graph,
  title={Graph neural networks in Alzheimer's disease diagnosis: a review of unimodal and multimodal advances},
  author={Ali, Shahzad and Piana, Michele and Pardini, Matteo and Garbarino, Sara},
  journal={Front. in Neurosci.},
  volume={19},
  pages={1623141},
  year={2025},
  publisher={Frontiers Media SA},
  doi={10.3389/fnins.2025.1623141}
}

@article{bevilacqua2023radiomics,
  title={Radiomics and artificial intelligence for the diagnosis and monitoring of Alzheimer’s disease: a systematic review of studies in the field},
  author={Bevilacqua, Roberta and Barbarossa, Federico and Fantechi, Lorenzo and Fornarelli, Daniela and Paci, Enrico and Bolognini, Silvia and Giammarchi, Cinzia and Lattanzio, Fabrizia and Paciaroni, Lucia and Riccardi, Giovanni Renato and others},
  journal={J. Clin. Med.},
  volume={12},
  number={16},
  pages={5432},
  year={2023},
  publisher={MDPI},
  doi={10.3390/jcm12165432}
}



\end{document}